\documentclass[letterpaper]{article} 
\usepackage{aaai24}  

\usepackage{times}  
\usepackage{helvet}  
\usepackage{courier}  
\usepackage[hyphens]{url}  
\usepackage{graphicx} 
\usepackage{natbib}  
\usepackage{caption} 
\usepackage{amsmath} 
\usepackage[colorinlistoftodos]{todonotes}
\usepackage{subcaption}
\usepackage{array}

\usepackage{algorithm}
\usepackage{algorithmic}

\usepackage{newfloat}
\usepackage{listings}
\DeclareCaptionStyle{ruled}{labelfont=normalfont,labelsep=colon,strut=off} 
\floatstyle{ruled}
\newfloat{listing}{tb}{lst}{}
\floatname{listing}{Listing}
\title{Text-to-Level Diffusion Models With Various Text Encoders for Super Mario Bros}
\author{
    Jacob Schrum,
    Olivia Kilday,
    Emilio Salas,
    Bess Hagan,
    Reid Williams
}
\affiliations{
    Southwestern University\\
    Georgetown, TX, USA\\
    \{schrum2,kildayl,salas2,haganb,williamsr\}@southwestern.edu
}

\begin{document}

\maketitle

\begin{abstract}
Recent research shows how diffusion models can unconditionally generate tile-based game levels, but use of diffusion models for text-to-level generation is underexplored. There are practical considerations for creating a usable model: caption/level pairs are needed, as is a text embedding model, and a way of generating entire playable levels, rather than individual scenes. We present strategies to automatically assign descriptive captions to an existing dataset, and train diffusion models using both pretrained text encoders and simple transformer models trained from scratch. Captions are automatically assigned to generated scenes so that the degree of overlap between input and output captions can be compared. We also assess the diversity and playability of the resulting level scenes. Results are compared with an unconditional diffusion model and a generative adversarial network, as well as the text-to-level approaches Five-Dollar Model and MarioGPT. Notably, the best diffusion model uses a simple transformer model for text embedding, and takes less time to train than diffusion models employing more complex text encoders, indicating that reliance on larger language models is not necessary. We also present a GUI allowing designers to construct long levels from model-generated scenes.
\end{abstract}

\section{Introduction}


Modern generative AI models use natural language input to create outputs in various modalities, including text, image, sound, and video. 
These technologies have also been applied to Procedural Content Generation (PCG), 
``the algorithmic creation of game content with limited or indirect user input'' \cite{shaker2016procedural}. The use of generative AI classifies these methods as PCG via Machine Learning (PCGML) \cite{summerville:tog2018}.


Many models have been used to generate levels for Super Mario Bros.\ and other tile-based games, including
Long Short-Term Memory networks \cite{summerville2016super}, Generative Adversarial Networks (GANs)~\cite{volz:gecco2018}, Variational Autoencoders (VAEs) \cite{thakkar:cog19},
Large Language Models (LLMs)~\cite{sudhakaran:nips23}, the Five-Dollar Model (FDM)~\cite{merino:aaai23}, and diffusion models~\cite{lee:mva23}. LLMs and FDM are text-guided, whereas diffusion models can be trained unconditionally or with text guidance. 
Although the use of text guidance in diffusion models is common in popular models like Stable Diffusion~\cite{rombach:cvpr2022SD}, text-guidance seems underexplored in the realm of tile-based game level generation, which is the focus of this paper.

Though it is no surprise that diffusion models can be used for this purpose, there are still many practical considerations in training a working model, including procuring a dataset of adequately descriptive captions, selecting a text embedding model to pair with diffusion, and creating levels of the desired size with the finished model. These issues are explored in this paper. Specifically, our contributions are:

\begin{enumerate}
    \item A method for automatically assigning captions to Mario level scenes that could be generalized to other domains given sufficient expert knowledge.
    \item A method of assessing the quality of text-conditioned generation that depends on the ability to automatically assign captions to scenes.
    \item A demonstration of how to use various types of text embedding models, both pretrained and trained from scratch, in a text-to-level diffusion pipeline.
    \item A comparison of various text embedding approaches in terms of adherence to input prompts, training time, diversity, and playability, which ultimately concludes that a simple transformer model with a limited vocabulary results in the best diffusion models.
    \item A mixed-initiative GUI that can be used to combine model-generated scenes into larger levels.
\end{enumerate}

\section{Related Work}



Many generative PCGML models exist.
Relevant work is split into 
unconditional models (no language input)
and text conditional models (using natural language).


\subsection{Unconditional Models}

Early work in PCGML used models like Long Short-Term Memory networks \cite{summerville2016super} to generate Mario levels. A survey of other early PCGML approaches came out in 2018 \cite{summerville:tog2018}

That same year, Generative Adversarial Networks (GANs) \cite{goodfellow2014generative} for level generation were introduced \cite{volz:gecco2018}. This work also applied latent variable evolution \cite{bontrager2018deep} to find scenes with desired properties for Mario. Use of GANs for level generation was quickly expanded upon. 
GANs were combined with Graph Grammars \cite{gutierrez2020zeldagan} and interactive evolution \cite{schrum2020interactive} to generate Zelda levels. Compositional Pattern Producing Networks were used to combine GAN-generated level scenes into global patterns \cite{schrum:gecco2020cppn2gan, schrum:tog2023}.
Mixed Integer Programming was used to repair levels discovered by latent quality diversity evolution \cite{fontaine:rss2021}. The ability to generate samples from a single input was explored in both Mario \cite{awiszus:aiide2020} and Minecraft \cite{awiszus:cog2021worldgan}. Levels for multiple games were generated by a single GAN trained to induce a common latent space on data from multiple games \cite{kumaran:aiide2020}.

The concept of searching a latent space to generate levels was also explored with Variational Auto-Encoders (VAEs) \cite{kingma:iclr13}. The earliest application of this was to Lode Runner \cite{thakkar:cog19}. Later work showed how to blend concepts across multiple games \cite{sarkar2020conditional}, and used VAEs for latent quality diversity evolution \cite{sarkar2021generating}.

Recently, diffusion models \cite{yang2023diffusionsurvey} have risen to prominence. They generate content via an iterative denoising process using a convolutional UNet. Diffusion models predict the presence of noise in a noisy image, so that said noise can be removed to produce a clean image. Trained models start with pure noise and derive quality output from it.
A popular example is Stable Diffusion \cite{rombach:cvpr2022SD}, which adds text conditioning to the UNet and combines it with a VAE so that diffusion is performed in a compressed latent space rather than at the scale of the full image.

Stable Diffusion was the basis of research in the game Doom showing how diffusion models can function as semi-playable game engines \cite{valevski2024diffusionmodelsrealtimegame}. The model was trained to predict the next screen frame conditioned on actions taken by a Reinforcement Learning agent (actions replaced text embeddings). A similar approach was used to simulate playing Super Mario Bros \cite{virtuals2024videogame}.

Though impressive, these models try to reproduce the game experience rather than generate new content, but there are recent examples of generating levels with diffusion. Unconditional diffusion models can indeed generate convincing Mario level scenes when trained on scenes from the original game \cite{lee:mva23}. \citeauthor{dai:aaai24}~(\citeyear{dai:aaai24}) took individual Mario/Minecraft levels and used an unconditional diffusion model to generate new levels at different scales that share the  distribution of elements from that one sample.

These methods produce playable levels, but lack control from text guidance. This is why evolution has often been combined with unconditional models to produce desired results, but defining a fitness function is generally more challenging than describing what is desired, so the next section describes PCGML approaches guided by text inputs.

\subsection{Text-Conditional Models}

Despite the frequent association of diffusion models with text guidance (e.g.\ Stable Diffusion), there is not much work applying text conditioning to diffusion models for level generation. An exception is recent work on Text-to-game-Map (T2M) models trained as part of the Moonshine system \cite{nie:aaai2025}, though the primary focus of Moonshine is the generation of synthetic captions by an LLM for the sake of training T2M models. The diffusion model from Moonshine relies on a model which we refer to as \texttt{GTE} below, as it is one approach to text embedding that we apply.

Text-to-Level approaches not based on diffusion also appear in the literature. Another model in the Moonshine paper is the Five-Dollar Model (FDM)~\cite{merino:aaai23}, a feed-forward model whose name emphasizes its minimal computational requirements. Previous FDM results indicate it is useful despite its simplicity, but it struggles with overfitting and lack of diversity in outputs.

Level generation with variants of the Large Language Model (LLM) GPT2 from OpenAI \cite{radford2019openAIGPT2} has been demonstrated in both Sokoban \cite{todd:fdg23} and Super Mario Bros \cite{sudhakaran:nips23}. We compare against this publicly available MarioGPT model below, though the complexity of the text prompts it understands is less ambitious than what our models are capable of.


\section{Methods}
\label{sec:methods}

We outline how training scenes are collected and combined with generated captions before training text embedding models, and then text-to-level diffusion models.

\subsection{Training Data}
\label{sec:data}

Full levels are from Super Mario Bros.\ 
and the Japanese Super Mario Bros.\ 2, 
a.k.a.\ \emph{The Lost Levels}, which was not initially released outside of Japan. This data comes from the Video Game Level Corpus (VGLC) \cite{summerville:vglc2016}, a repository with data for several games. Despite being widely used~\cite{volz:gecco2018,sudhakaran:nips23,lee:mva23}, the data has numerous errors and omissions, so we use our own manually cleaned version that is closer to data from the real 
games\footnote{\url{https://github.com/schrum2/TheVGLC}}, 
and adds back some missing levels. 
However, we retain the limitation 
of representing enemies with a single symbol interpreted as a Goomba. We thus have 13 tile types.


As in previous works~\cite{volz:gecco2018,lee:mva23}, characters for each tile correspond to integers that are one-hot encoded. This approach has proven sufficient for us and others, though vector-based block/tile embeddings have also been used~\cite{awiszus:cog2021worldgan, dai:aaai24}. Such an approach could be useful, but is not explored in this paper.

To extract data, a window slides over each level one tile at a time. Mario levels are 14 tiles high, but because the architectural components of our diffusion model are easier to define when input sizes are powers of 2, we pad the tops of levels 
to create $16 \times 16$ samples. 

Creating descriptive captions for each scene is more complicated. The Moonshine system~\cite{nie:aaai2025} mentioned previously uses LLMs to create suitable captions for level data, but we use a deterministic approach. Concepts from Mario scenes are manually defined, and scenes are scanned for the presence and quantity of these concepts, resulting in up to one phrase ending in a period for each concept. 
The full list of concepts and how they are defined is here:


\begin{itemize}
\item Floor: Blocks on the bottom row. Can have gaps, or be a void with small floor chunks.
\item Ceiling: Blocks in fourth row filling at least half of the row. Can have gaps. 
\item Pipe: Four correctly arranged tiles of a pipe with neck tiles extended to a solid base or the bottom of the screen. 
\item Upside down pipe: Pipe with opening at the bottom and neck that
extends to a solid top or the top of the screen.
\item Coin Line: Adjacent coins in the same row. 
\item Coin: Coin tiles. Includes coins in lines. 
\item Cannon: Cannon tiles. 
\item Question Block: Both types of question block tiles. 
\item Enemy: Enemy tiles. 
\item Platform: Adjacent solid tiles in the same row, with the rows above and below being empty/passable. 
\item Tower: Collection of contiguous blocks with a width less than three and a height of at least three.
\item Ascending Staircase: Solid tiles with empty space above where height increases by one for each move to the right. Sequence is at least three columns wide.  
\item Descending Staircase: Like ascending staircase, but height decreases by one while moving right. 
\item Rectangular Cluster: Flood-filled rectangular cluster of contiguous solid blocks. Flood fill excludes previously identified structures. 
\item Irregular Cluster: Remaining flood-filled clusters of at least three contiguous blocks not captured earlier. 
\item Loose Block: Solid blocks not captured earlier. 
\end{itemize}

\begin{figure}[t]
\centering

    \begin{subfigure}[t]{0.22\textwidth}
        \centering
        \includegraphics[width=\linewidth]{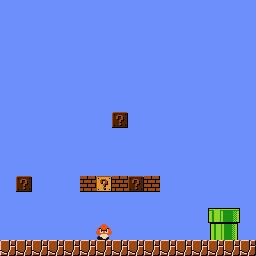} 
        \caption{Real Scene}
        \label{fig:real_scene}
    \end{subfigure}%
    \hfill
    \begin{subfigure}[t]{0.22\textwidth}
        \centering
        \includegraphics[width=\linewidth]{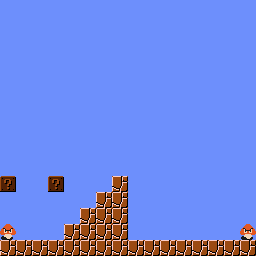} 
        \caption{Generated Scene}
        \label{fig:generated_scene}
    \end{subfigure}


\caption{Real and Generated Level Scenes.
(\subref{fig:real_scene}) A scene from the data. Its captions are: \texttt{regular}: ``full floor. one enemy. a few question blocks. one platform. one pipe.'' \texttt{absence}: ``full floor. no ceiling. one enemy. a few question blocks. no cannons. no coins. no coin lines. one platform. no ascending staircases. no descending staircases. one pipe. no upside down pipes. no towers. no rectangular block clusters. no irregular block clusters. no loose blocks.'' \texttt{negative}: ``ceiling. cannon. coin. coin line. ascending staircase. descending staircase. upside down pipe. tower. rectangular block cluster. irregular block cluster. loose block.''
(\subref{fig:generated_scene}) Model-generated scene. Input prompt: ``floor with one gap. a few platforms. a few enemies. a few coins. one coin line. a few towers. one ascending staircase. a few question blocks.'' Actual caption: ``full floor. two enemies. one ascending staircase. two question blocks.''
Resulting caption adherence score: 0.478.}
\label{fig:scenes_and_captions}
\end{figure}


Most concepts include a quantity: ``one'', ``two'', ``a few'' (3-4), ``several'' (5-9), or ``many'' (10 or more). The floor concept distinguishes between ``full floor'' and one with some number of gaps. If over half of the floor is missing, it is a ``giant gap'' with some number of ``chunks of floor'', though some levels have no floor. Similarly, a ceiling is either ``full'' or has some number of gaps. 

This captioning style is the \texttt{regular} approach. However, we also consider \texttt{absence} captions, in which every concept missing from a scene is explicitly mentioned, as in ``no floor.'' The \texttt{absence} captions always have the same number of phrases, whereas the number in \texttt{regular} captions varies. Examples of each captioning approach are in Figure \ref{fig:real_scene}.
To encourage flexibility in using the models, the order of the phrases in training captions is randomized.


We can also assign captions to artificial scenes output by our models, which is useful for assessing model controllability later.
When assigning captions to model output, two additional concepts are potentially present:

\begin{itemize}
\item Broken Pipes: Portions of a pipe that lack one of the four required tiles, or place them inappropriately.
\item Broken Cannons: When a cannon support tile appears without a cannon tile on top.
\end{itemize}

\noindent These concepts are never present in training data.

\subsection{Text Embedding Models}

The text-conditional models in Related Work depend on pretrained language models to embed text input for the level generator. Leveraging existing models allows for open-ended text input. However, game environments are constrained, so a large vocabulary is not necessary. Therefore, we train a simple model from scratch on limited vocabulary.



The simple architecture we use is a standard transformer encoder that learns token embeddings of length 128. Full details are in the appendix, but we allow multiple transformer encoder layers with multi-headed self-attention.
During training, the model is given a sequence of token IDs, and encodes them with an embedding layer. These embeddings are combined with sinusoidal positional encodings and passed through the transformer layers where the attention mechanism enriches the embedded representations with surrounding context.
For the sake of training, these embeddings are passed through a final linear layer that outputs logits for each token at each position. Masked Language Modeling is used during training, so we refer to this model as \texttt{MLM}. This means that some input tokens are probabilistically replaced with a special MASK token, but the model must predict the correct tokens from surrounding context. The result is token embeddings that capture semantic information about their context in sentences from the training data.

\texttt{MLM} is a small model trained on a small dataset with a small vocabulary, so it trains quickly and is effective at modeling the restricted grammar in our captions. However, \texttt{MLM} cannot tolerate tokens not present in its training data. In contrast, pretrained language models have been used in previous text-to-level systems, and \emph{can} accept arbitrary tokens outside our limited vocabulary.
The original FDM paper~\cite{merino:aaai23} used the sentence transformer \texttt{multi\allowbreak-qa\allowbreak-MiniLM\allowbreak-L6\allowbreak-cos\allowbreak-v1} (\texttt{MiniLM}). \texttt{MiniLM} maps whole sentences to vectors of length 384, and was designed for semantic search. Later research with Moonshine~\cite{nie:aaai2025} combined FDM and a diffusion model with \texttt{gte\allowbreak-large\allowbreak-en\allowbreak-v1.5} (\texttt{GTE}) \cite{zhang-etal-2024-mgte}. \texttt{GTE} embeds entire documents into vectors of length 1024. 
Although these models allow arbitrary tokens, such tokens have little meaning in Mario, and even familiar tokens are used differently in our captions than in natural language.
Therefore, further fine-tuning of these sentence transformers could be useful, 
but this is left for future work.


An alternative approach that is explored takes advantage of the form of our captions: collections of period-separated phrases. The default approach embeds each caption as a \texttt{single} vector, but these phrases can be embedded individually to create \texttt{multiple} vectors to provide to the diffusion model. Both approaches are applied 
in our experiments.

\subsection{Diffusion Model}



The diffusion model is a conditional UNet with 13 in/out channels: one per tile type. 
It has three convolutional down-sampling stages and three up-sampling stages.
Each contains residual blocks with SiLU activations and skip connections to preserve spatial information and avoid disappearing gradients, as well as cross-attention to allow text-embedding input from the language models described in the previous section. The 13 channels project to 128 channels, then 256, then 512 at the bottleneck before reversing the sequence.
An unconditional model can be easily made by removing the cross-attention, and is done for the sake of comparison with an approach similar to that of \citeauthor{lee:mva23} (\citeyear{lee:mva23}).



We use classifier-free guidance, meaning the model is trained to handle both conditional and unconditional inputs, allowing stronger guidance at inference by interpolating between the two. Specifically, the text conditional model is trained on each sample using both text embeddings and an empty embedding vector. Effectively doubling the samples increases training time in comparison with an unconditional model. We can use \texttt{regular} or \texttt{absence} captions, but there is also a third option. Instead of indicating the absence of items in the caption, we can train with distinct negative prompts. Because all possible concepts are known, we can take each \texttt{regular} caption and make a separate negative prompt listing all concepts that are absent. Now for each sample, a third copy is paired with the negative prompt for negative guidance. This \texttt{negative} caption approach takes even longer to train since it triples the number of samples. An example scene with all captions is in Figure \ref{fig:real_scene}.


The diffusion model is given noisy one-hot encoded level scenes as input, and tries to predict the noise that would need to be removed to get the original input. 
The loss function is a weighted sum of mean squared error (MSE) and categorical cross-entropy (reconstruction loss), as in \citeauthor{lee:mva23} (\citeyear{lee:mva23}). Detailed diffusion diagram in appendix.






\section{Experiments}

We train numerous 
models, and evaluate them with various metrics. All models were trained on different lab machines sharing the same hardware: 
Alienware, 13th Gen Intel\textsuperscript{\textregistered} Core\texttrademark~i9-13900F with 24 cores at base speed of 2.0GHz, 32 GB RAM, 
NVIDIA GeForce RTX 4060 with 8 GB dedicated VRAM and 15.9 GB shared VRAM. 
These are reasonably powerful gaming PCs.
Detailed parameter settings and additional results are available in an appendix at \url{https://people.southwestern.edu/~schrum2/mario.html} and
source code for recreating our results along with selected models are available at \url{https://github.com/schrum2/MarioDiffusion}.

\subsection{Dataset Preparation}


We use a 90/5/5-split of the 7,687 samples from the two Mario games to get training, validation, and test sets of sizes 6,918, 384, and 385. Care is taken to assure that all three datasets contain representation of all possible concepts.
For training both text and diffusion models, data is augmented via random shuffling of phrases within captions. Validation data is used during training to determine the best model to keep, and test data is used for evaluation after training. We also make a set of 100 randomly generated captions not present in the original data for further testing. Each contains random phrases for 1 to 10 randomly selected topics.

\subsection{Training Text Embedding Models}

We train a separate \texttt{MLM} text encoder for each diffusion model that uses one. Models using \texttt{regular} and \texttt{absence} captions are trained with those caption types, though \texttt{MLM} models for \texttt{negative} captions also use \texttt{regular} data. Each model is trained for 300 epochs using AdamW with cross-entropy loss, but the lowest validation loss is logged every epoch, and the model with the best validation loss is kept as the final model.

\subsection{Training Text-Conditioned Diffusion Models}

For each text embedding approach and caption style, we train as many models as practical given compute costs. 
We train models with different random seeds for each caption style: 10 \texttt{MLM},
10 \texttt{MiniLM\allowbreak-single}, 
5 \texttt{MiniLM\allowbreak-multiple}, 
5 \texttt{GTE\allowbreak-single}, 
and 1 \texttt{GTE\allowbreak-multiple}.


Models are trained with AdamW for 500 epochs
using a cosine learning rate schedule and a warm-up period.
To prevent overfitting, the caption adherence score defined later in
Equation \ref{eqn:caption_score} is computed every 20 epochs across all validation captions,
so the final model is whichever
one had the best average $\text{c-score}$.
The average $\text{c-score}$ is a better measure of model performance than the usual validation loss.

\subsection{Comparison Models}

For comparison, we train several other models.
We train 30 unconditional diffusion models for 500 epochs in a manner similar to \citeauthor{lee:mva23} (\citeyear{lee:mva23}). Since there are no captions, best model is determined by the lowest validation loss.
We also train 30 Wasserstein GANs (WGANs) following the methodology of \citeauthor{volz:gecco2018} (\citeyear{volz:gecco2018}), meaning we train for 5,000 epochs and the final model is from the final epoch.

Five-Dollar Models (FDMs) \cite{merino:aaai23} 
are trained using \texttt{MiniLM} and \texttt{GTE} as embedding models
with \texttt{regular} and \texttt{absence} captions (30 models per combination). As input, FDM
takes a sentence embedding vector and a noise vector of length 5.
As a text-conditioned model, FDM checks caption adherence score on validation data every 10 epochs to determine
the best final model. However, our experience confirms the observation from
previous work that FDM is prone to overfitting, so it is only trained for 100 epochs.
Although the caption adherence scores for FDM show a general upward trend (with occasional dips),
validation loss increases after an initial dip early in training.



We also compare against MarioGPT's publicly available model \cite{sudhakaran:nips23}, but do not train our own version. MarioGPT's repertoire of training captions is comparatively limited, based on only 96 combinations (barring the use of arbitrary integer quantities). We use each caption to generate a level 128 blocks long and slice each into 8 scenes, from which 100 scenes are sampled.


\subsection{Measuring Performance}


\subsubsection{Caption Adherence Score}
We focus on the ability of text-to-level models
to produce scenes matching their input
prompts. As indicated earlier, 
we can automatically assign captions to
any level scene, including output from trained models.
Output captions can be compared to their corresponding input
prompts to define a caption score ($\text{c-score}$):

\begin{equation}
\label{eqn:caption_score}
\text{c-score}(p,c) = \frac{\sum_{t \in T} \text{match}(\text{phrase}(t,p),\text{phrase}(t,c))}{|T|} 
\end{equation}



\begin{equation}
\text{match}(p_t,c_t) = 
\begin{cases}
        1.0     & \text{for } p_t = c_t\\
        1.0 - \frac{|\text{qu}(p_t) - \text{qu}(c_t)|}{|Q| - 1} & \text{for } \text{co}(p_t) \land \text{co}(c_t) \\
        0.1  & \text{for } p_t \neq \emptyset \land c_t \neq \emptyset \\
        -1.0 & \text{otherwise}
\end{cases}
\end{equation}


\noindent $p$ is a prompt. $c$ is a caption describing
the scene produced from $p$.
$T$ is the set of caption concepts for Mario levels.
$\text{phrase}(t,s)$ returns the phrase in $s$
associated with concept $t$, or $\emptyset$ if there is no such phrase or it starts with ``no'' (possible for \texttt{absence} captions).
$\text{co}(s_t)$ is for \emph{countable}, and indicates whether $s_t$ describes a quantity (phrases like ``full floor'' do not). 
$\text{qu}(s_t)$
returns an integer that orders quantities: 
``one'' = 0 up to ``many'' = 4. 
$Q$ is the set of quantities. 


So, $\text{match}(p_t,c_t)$ takes two phrases about the same concept, and returns 1.0 if they are identical (first case), a value from 0.0 to 1.0 if there is a partial match (second and third cases), and -1.0 if one phrase indicates the concept is completely absent, but the other does not. If both phrases indicate that some quantity of the concept is present, then smaller differences in quantities result in higher results. 
The third case returns 0.1 when one phrase has a quantity and the other does not, which only happens when comparing a full floor or ceiling to one with some amount of empty space.


The $\text{c-score}(p,c)$ calculation is simply the average $\text{match}(p_t,c_t)$ score across all topics, which means it is scaled from the range -1.0 to 1.0.
 An example comparison and resulting $\text{c-score}$ are in Figure \ref{fig:generated_scene}.


\subsubsection{End Time and Best Time}

A model's training time is also relevant. We define \emph{End Time} as the time required to complete the final epoch, and \emph{Best Time} as the time required to 
reach the checkpoint with the best validation performance. 
\emph{Best Time} is relevant since early stopping could hopefully end training shortly after a model reaches its best performance.
Note that times for \texttt{MLM} diffusion models include the additional training time for the actual \texttt{MLM} text encoders.

\subsubsection{Average Minimum Edit Distance}

This metric measures the variety of levels
within a set. 
Levels are compared in terms of edit distance. 
The average minimum edit distance $\text{AMED}_{\text{self}}$ across a set of level scenes $S$ is

\begin{equation}
\text{AMED}_{\text{self}}(S) = \frac{\sum_{s \in S} \text{argmin}_{x \in (S - \{s\})} \text{dist}(s,x)}{|S|} 
\end{equation}


A high $\text{AMED}_{\text{self}}$ score means most levels are very different from each other. A low score means most levels are similar to other levels in the set. As a set of limited possibilities grows, closer neighbors are more likely to be found, so comparison of sets with equal sizes is important for fairness.


Average minimum edit distance can also be defined with respect to 
real game data to define $\text{AMED}_{\text{real}}$:

\begin{equation}
\text{AMED}_{\text{real}}(S) = \frac{\sum_{s \in S} \text{argmin}_{x \in R} \text{dist}(s,x)}{|S|} 
\end{equation}

\noindent $R$ is the set of 
real game samples.
High $\text{AMED}_{\text{real}}$ scores indicate 
large differences between generated and real levels. 


\subsubsection{Solvability}

A level scene is considered solvable if Robin Baumgarten's A* agent \cite{togelius:cig2010:marioAI} can beat it, though this widely used agent is not perfect \cite{sosvald:ictai2021marioastar}, and we have observed unusual failure cases. Furthermore, it is not always the agent's fault that a scene cannot be beaten.
Although complete levels are beatable, slicing them into $16 \times 16$ samples sometimes results in the loss of a platform or other element that is required to traverse the remainder of the scene. Only 7,160 of the 7,687 samples are solvable, approximately 93\%.

\begin{figure*}[t]
\centering

    \begin{subfigure}[t]{0.33\textwidth}
        \centering
        \includegraphics[width=\linewidth]{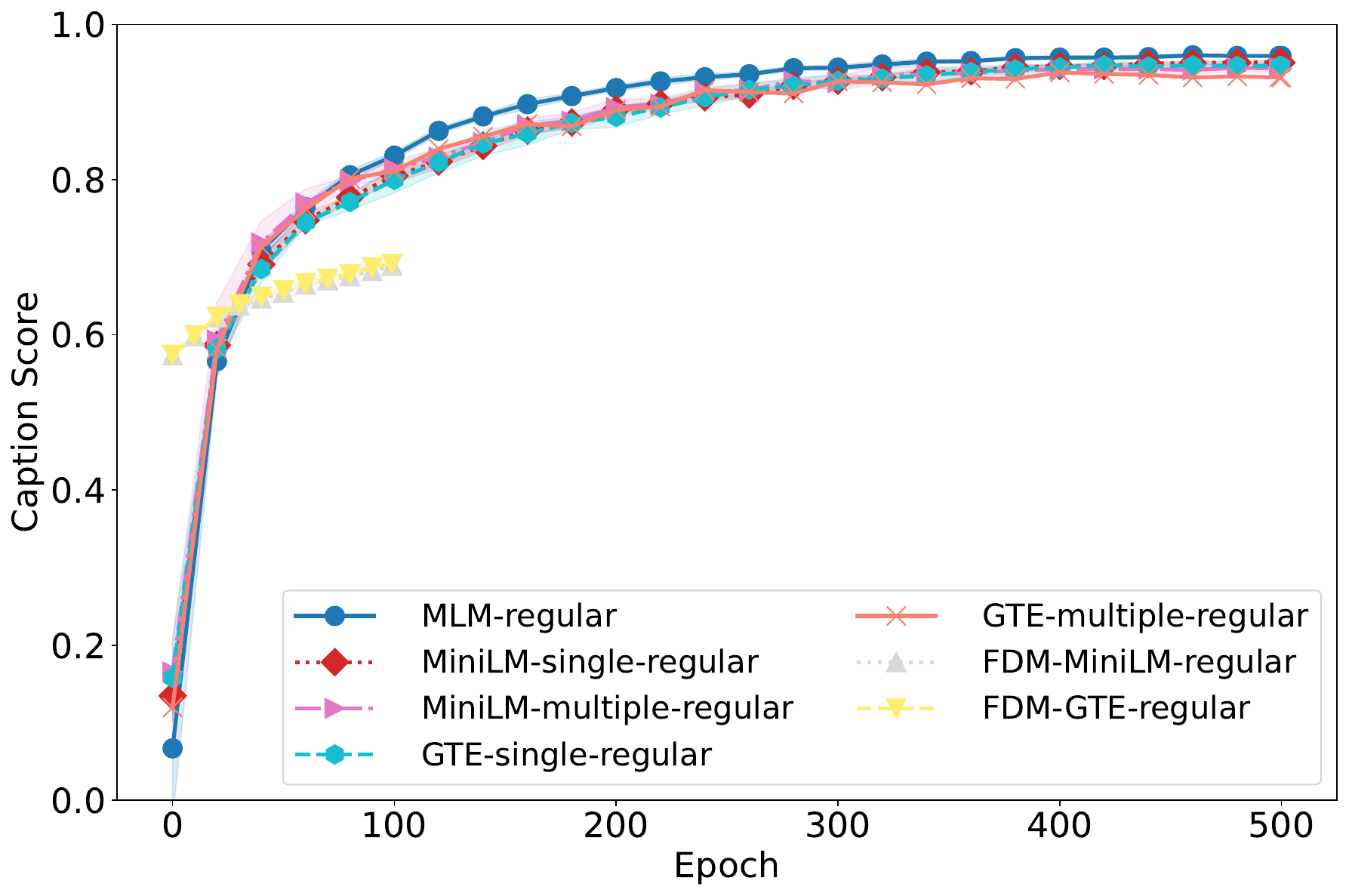} 
        \caption{\texttt{regular}}
        \label{fig:real_regular}
    \end{subfigure}%
    \hfill
    \begin{subfigure}[t]{0.33\textwidth}
        \centering
        \includegraphics[width=\linewidth]{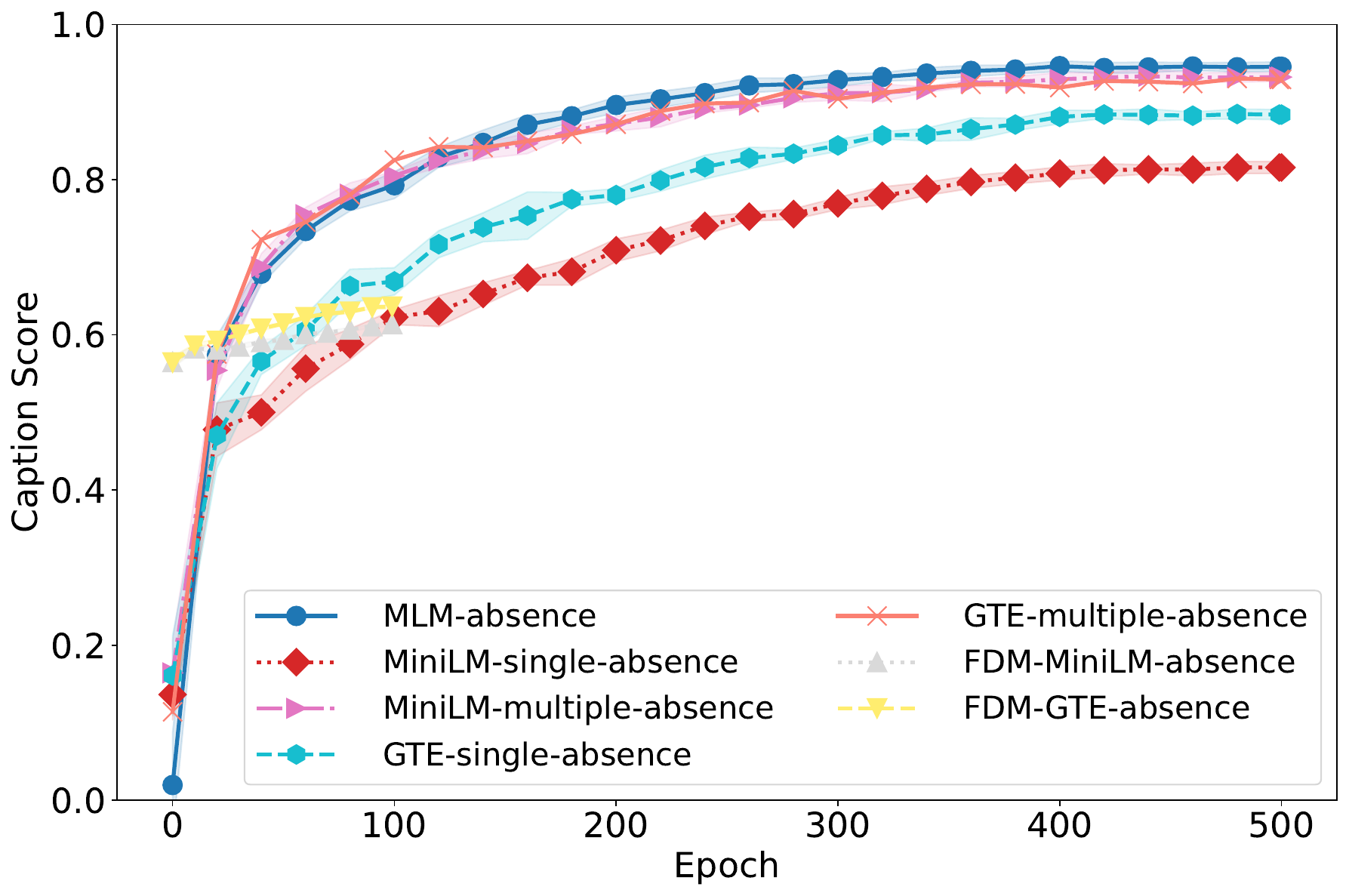} 
        \caption{\texttt{absence}}
        \label{fig:real_absence}
    \end{subfigure}
    \hfill
    \begin{subfigure}[t]{0.33\textwidth}
        \centering
        \includegraphics[width=\linewidth]{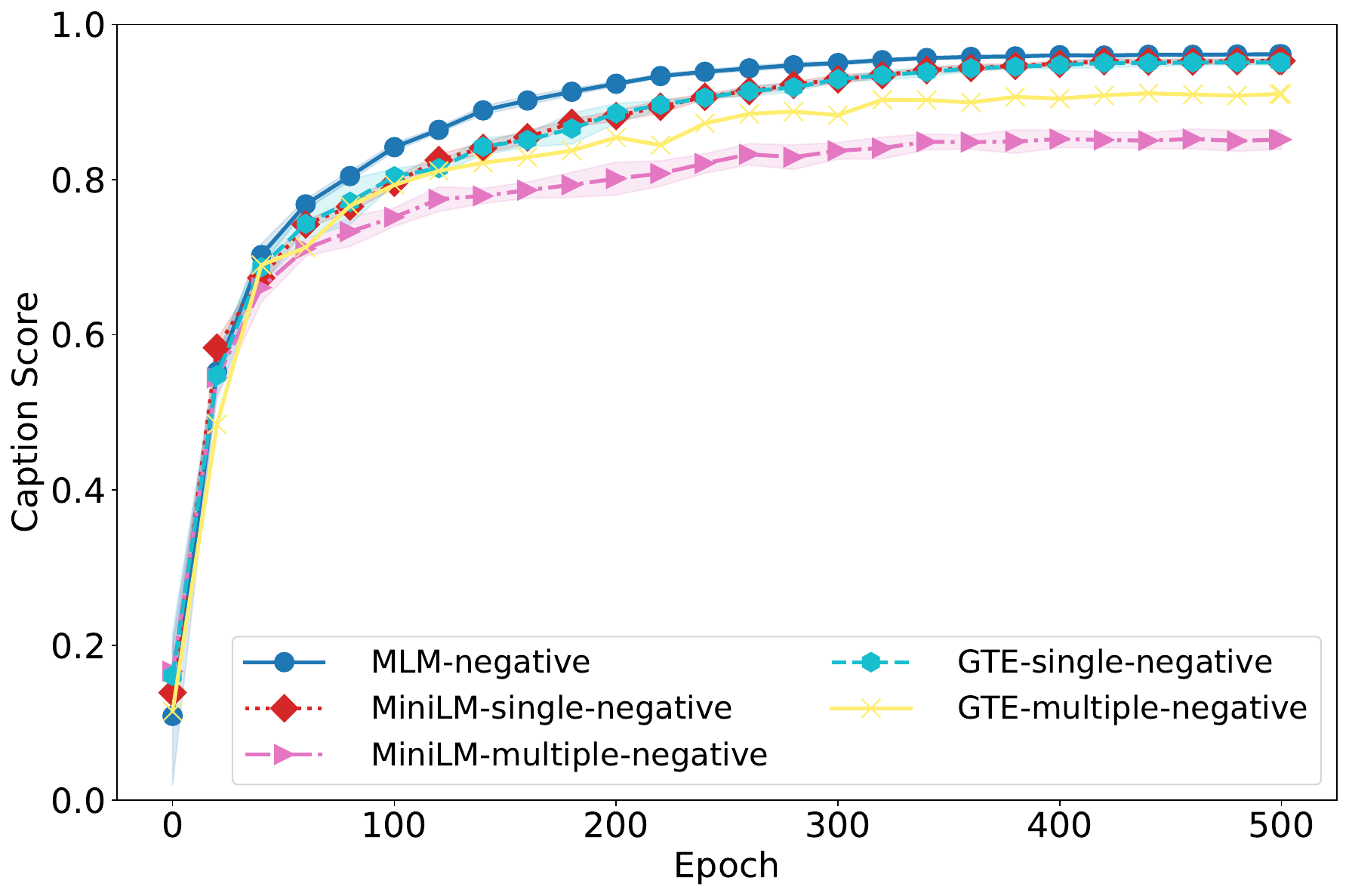} 
        \caption{\texttt{negative}}
        \label{fig:real_negative}
    \end{subfigure}

\caption{Average Caption Adherence Score by Epoch on Real Game Test Set Captions. The number of runs being averaged over varies by model type, though \texttt{GTE-multiple} data are based on a single run each. When there is more than one run, 95\% confidence intervals are shown, though they are often extremely narrow. Recall that FDM models are trained for a shorter period due to their overfitting problem.
(\subref{fig:real_regular}) \texttt{regular} models are all good, though \texttt{MLM} peaks faster. FDM starts high but lags behind.
(\subref{fig:real_absence}) \texttt{absence} results are more varied, with \texttt{GTE-single} and \texttt{MiniLM-single} performing poorly, and \text{MLM} once again on top. FDM results are flatter.
(\subref{fig:real_negative}) \texttt{negative} results are also varied, with \texttt{MLM} once again on top and poor performance from \texttt{MiniLM-multiple} and \texttt{GTE-multiple}.
FDM does not train with \texttt{negative} captions.}
\label{fig:caption_real}
\end{figure*}

\begin{figure*}[th]
\centering

    \begin{subfigure}[t]{0.33\textwidth}
        \centering
        \includegraphics[width=\linewidth]{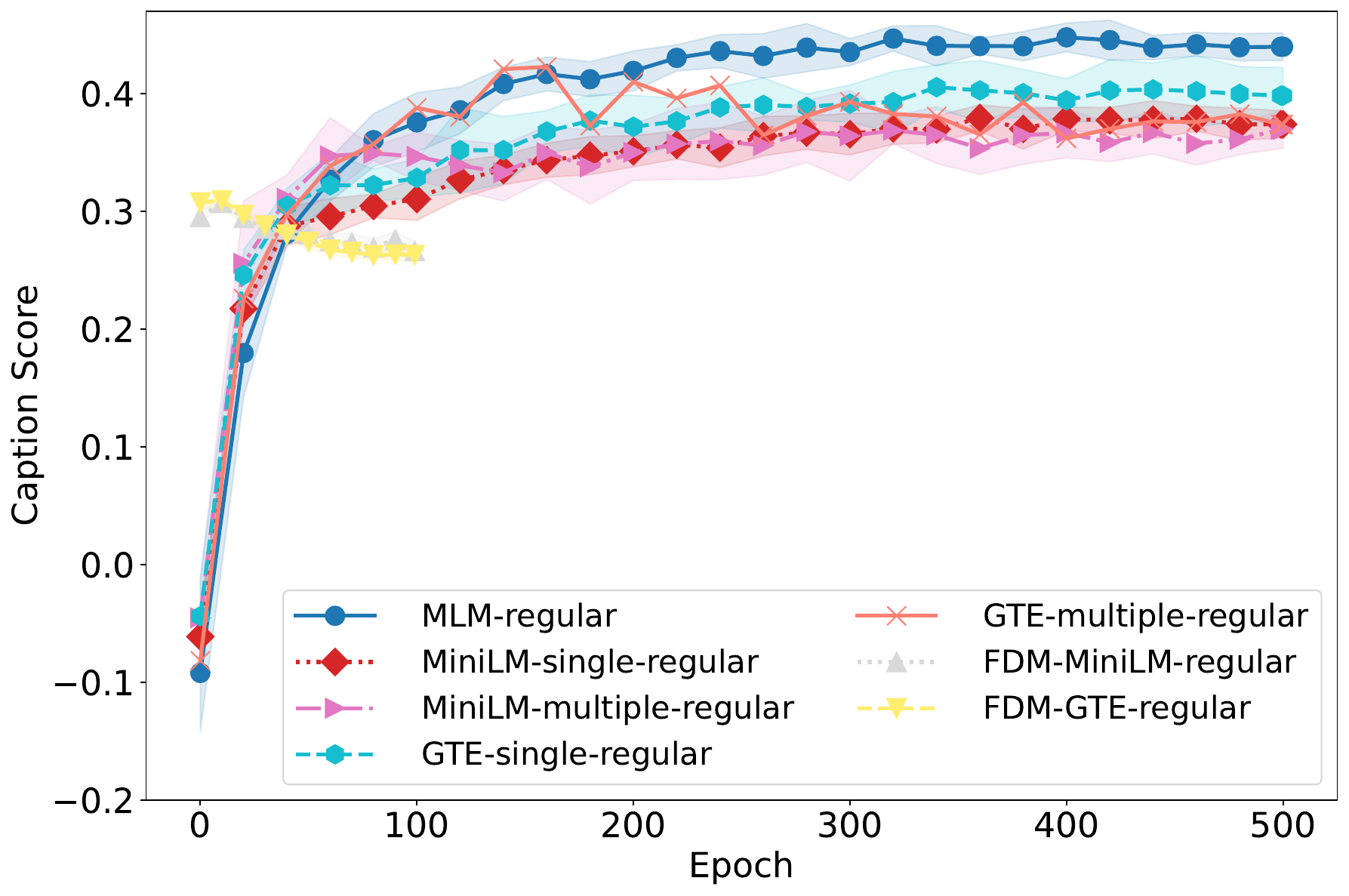} 
        \caption{\texttt{regular}}
        \label{fig:random_regular}
    \end{subfigure}%
    \hfill
    \begin{subfigure}[t]{0.33\textwidth}
        \centering
        \includegraphics[width=\linewidth]{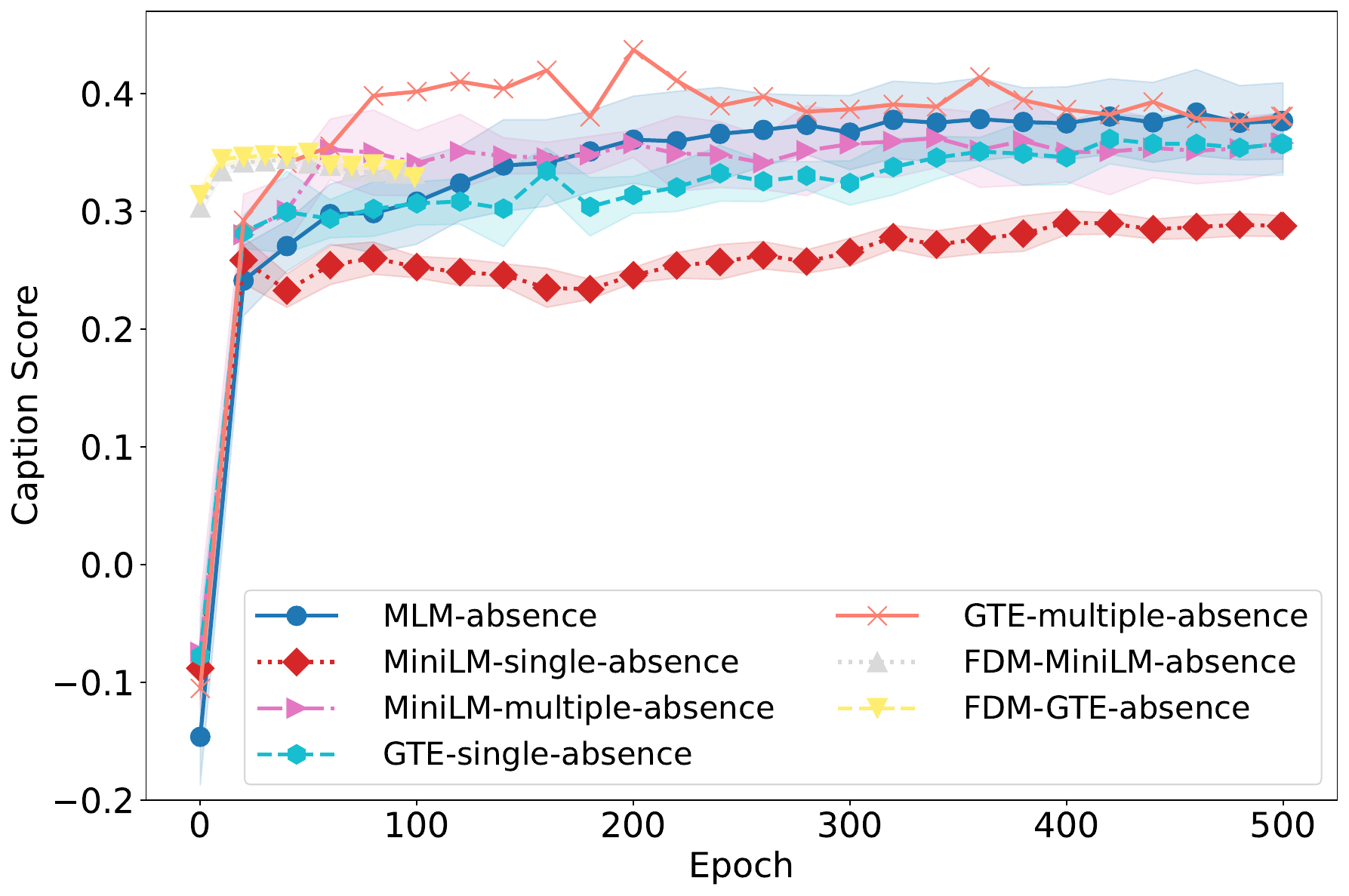} 
        \caption{\texttt{absence}}
        \label{fig:random_absence}
    \end{subfigure}
    \hfill
    \begin{subfigure}[t]{0.33\textwidth}
        \centering
        \includegraphics[width=\linewidth]{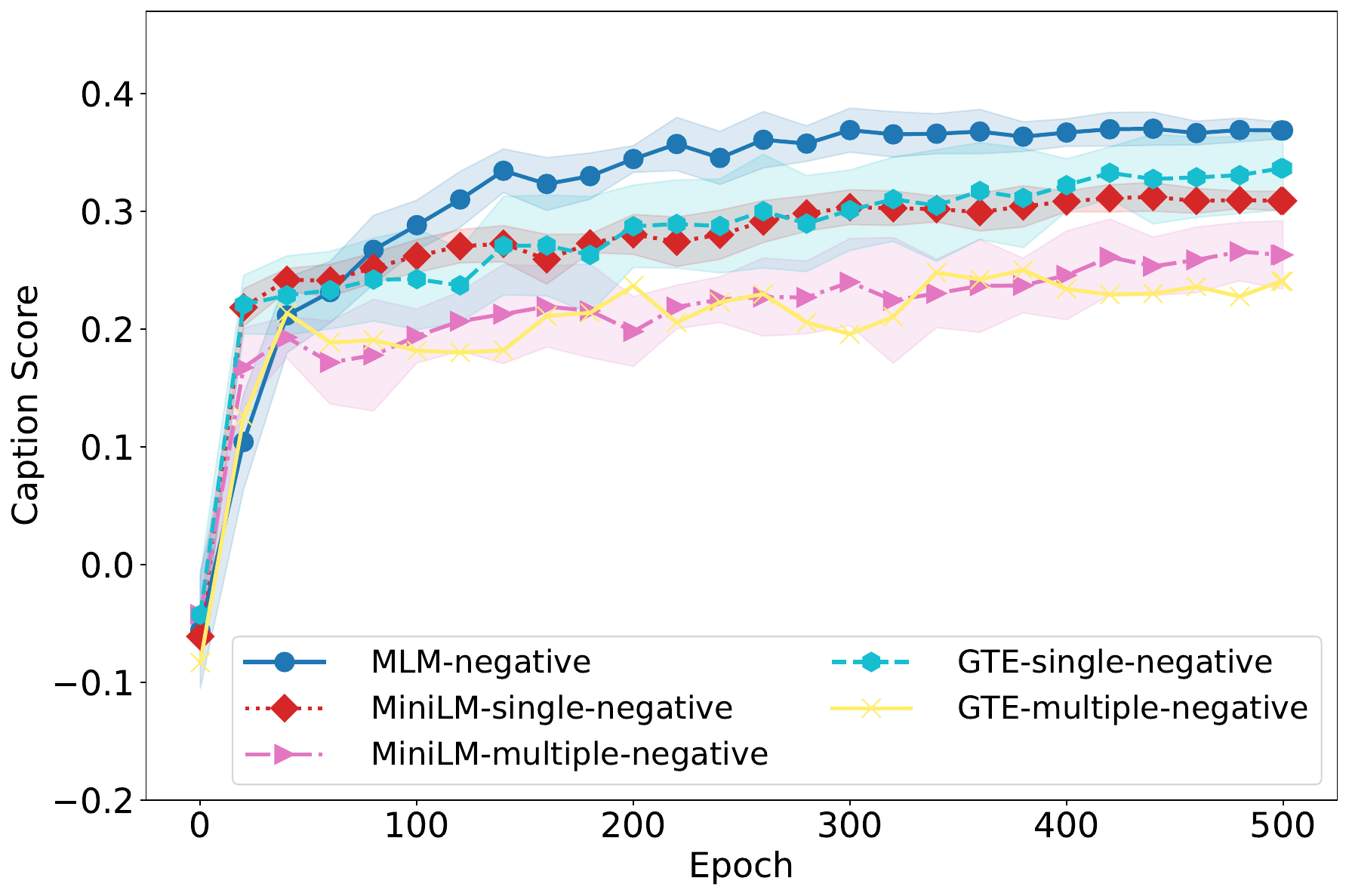} 
        \caption{\texttt{negative}}
        \label{fig:random_negative}
    \end{subfigure}

\caption{Average Caption Adherence Score by Epoch on Random Captions. The 95\% confidence intervals are more evident here.
(\subref{fig:random_regular}) \texttt{regular} models have the highest scores. \texttt{MLM} is the best among them. FDM performance plummets quickly and \texttt{GTE\allowbreak-multiple} is erratic.
(\subref{fig:random_absence}) \texttt{absence} data mostly has worse scores, though \texttt{GTE\allowbreak-multiple} has a spike before dipping to meet the other models. Only \texttt{MiniLM\allowbreak-single} is exceptionally low. FDM models maintain performance better in this case.
(\subref{fig:random_negative}) \texttt{negative} performance is the worst, though \texttt{MLM} is the best in this group. FDM does not train with \texttt{negative} captions.}
\label{fig:caption_random}
\end{figure*}

\subsubsection{Level Integrity}


In model output, tiles for pipes and cannons are sometimes arranged incorrectly.
The broken pipe issue was first recognized when generating Mario levels with GANs \cite{volz:gecco2018}. There are trivial ways to repair broken features or change the data encoding so that they cannot appear \cite{schrum:gecco2020cppn2gan}, but forcing the models to learn how to build such structures provides us another way to assess them.
Therefore, the percentage of generated scenes with broken pipes and cannons is reported. 


\section{Results}
\label{sec:results}



\subsubsection{Caption Adherence Score}

Caption adherence across the test set prompts from real game scenes is in Figure \ref{fig:caption_real}.
Most models earn scores above 0.9 within 500 epochs. Among text-conditioned diffusion models, the only exceptions are
\texttt{MiniLM\allowbreak-single\allowbreak-absence}, 
\texttt{GTE\allowbreak-single\allowbreak-absence}, 
\texttt{MiniLM\allowbreak-multiple\allowbreak-negative}, 
and \texttt{GTE\allowbreak-multiple\allowbreak-negative}. FDM models start high, but get stuck around 0.6 to 0.7 depending on the specific model.

Results across all real data are qualitatively similar (appendix).
In contrast, models have difficulty with 
random captions (Figure \ref{fig:caption_random}). The highest score is under 0.5, and is achieved by \texttt{MLM-regular}. 
Figure \ref{fig:generated_scene} shows an example scene generated by \texttt{MLM\allowbreak-regular}. 
Early in training, \texttt{GTE\allowbreak-multiple\allowbreak-regular} and \texttt{GTE\allowbreak-multiple\allowbreak-absence} reach the maximum score achieved by \texttt{MLM\allowbreak-regular} before dropping down. Overfitting by FDM is evident, as performance peaks and then drops within about 30 epochs. The worst scores are from \texttt{MiniLM\allowbreak-single\allowbreak-absence}, \texttt{MiniLM\allowbreak-multiple\allowbreak-negative}, and \texttt{GTE\allowbreak-multiple\allowbreak-negative}.
The way some scores drop indicate that random captions may have served better to determine the best final model than validation captions.


\begin{figure*}[t]
\centering

    \begin{subfigure}[t]{0.33\textwidth}
        \centering
        \includegraphics[width=\linewidth]{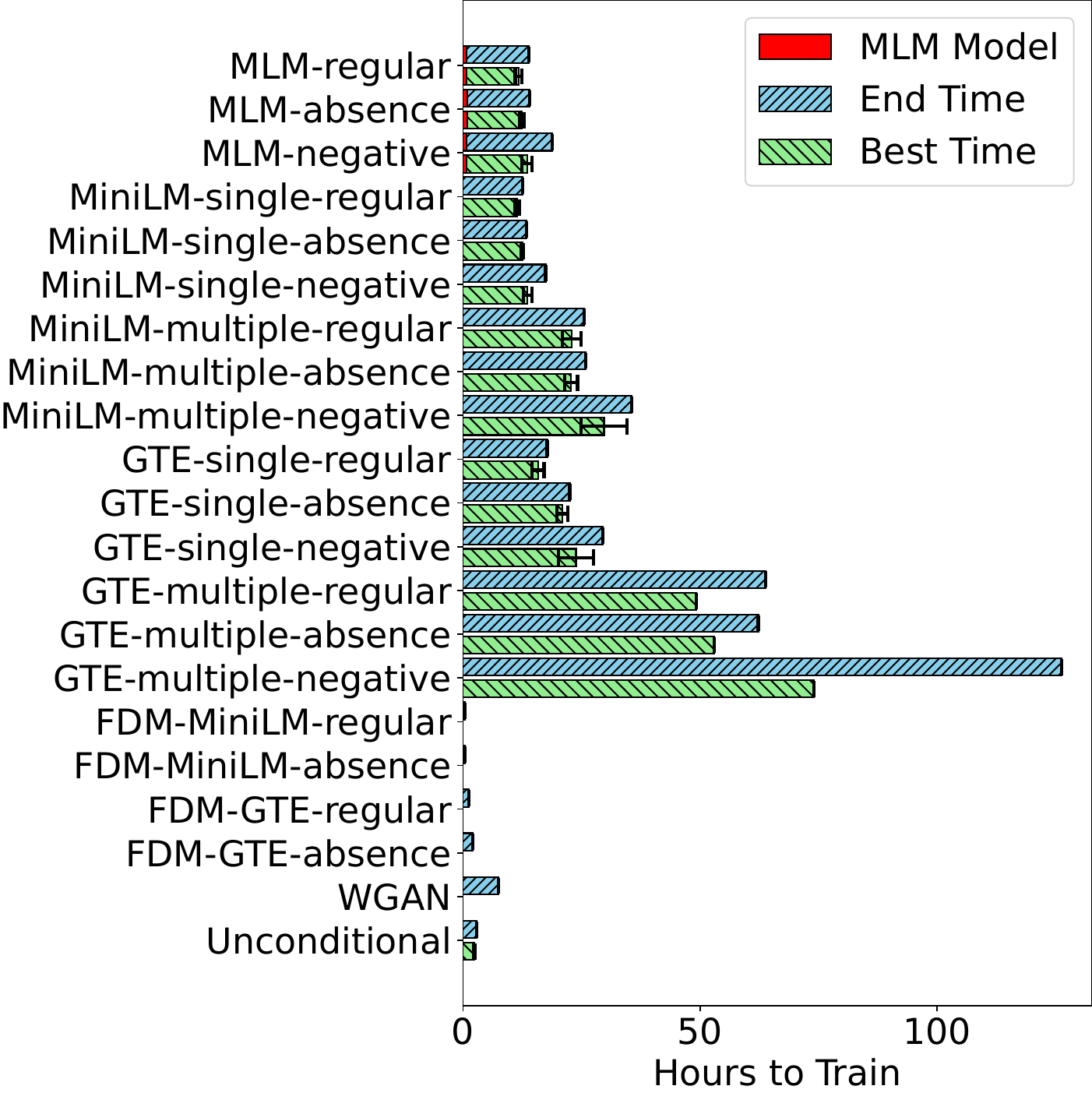} 
        \caption{Average End Times and Best Times}
        \label{fig:time_and_epoch}
    \end{subfigure}%
    \hfill
    \begin{subfigure}[t]{0.33\textwidth}
        \centering
        \includegraphics[width=\linewidth]{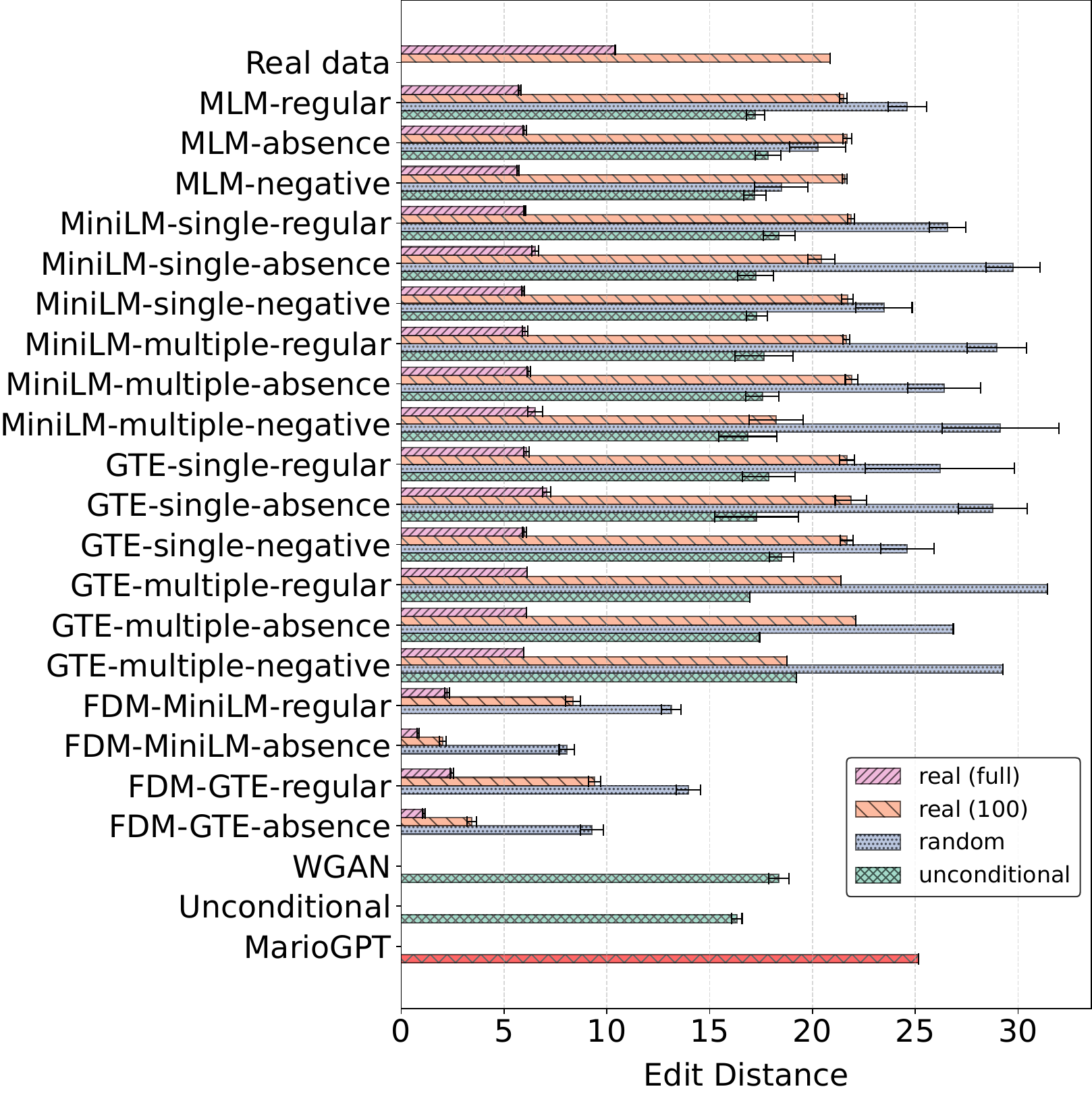} 
        \caption{$\text{AMED}_{\text{self}}$ on Model Outputs}
        \label{fig:AMED_self}
    \end{subfigure}%
    \hfill
    \begin{subfigure}[t]{0.33\textwidth}
        \centering
        \includegraphics[width=\linewidth]{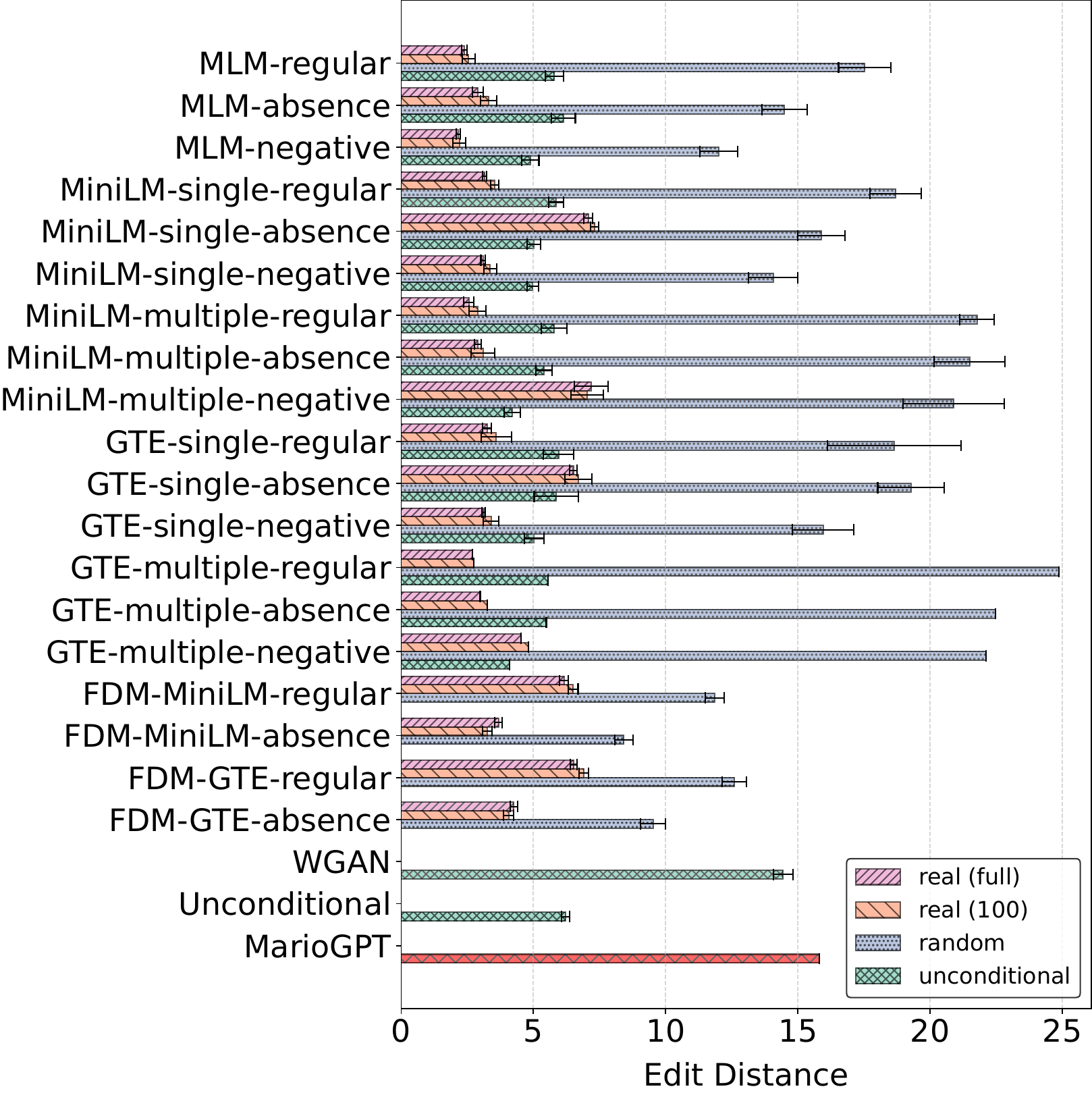} 
        \caption{$\text{AMED}_{\text{real}}$ on Model Outputs}
        \label{fig:AMED_real}
    \end{subfigure}

\caption{Bars have 95\% confidence intervals when multiple models were trained.
(\subref{fig:random_regular}) 
Time to train the text encoder for \texttt{MLM} diffusion models is shown in red on the left, but is barely visible. \emph{Best Time} is not shown for WGAN or FDM.
(\subref{fig:AMED_self}) $\text{AMED}_{\text{self}}$ compares the diversity of model-generated scenes. 
\emph{real (full)} is generated from all real data prompts.
\emph{real (100)} is from a selection of 100 real prompts.
\emph{random} prompts are distinct from real prompts.
\emph{unconditional} scenes had no prompts.
Diversity for Real data is also shown, and data from MarioGPT. WGAN and Unconditional cannot produce scenes from prompts. FDM has no \emph{unconditional} scenes. 
(\subref{fig:AMED_real}) $\text{AMED}_{\text{real}}$ compares model-generated scenes in terms of their distinctness from the real data.}
\label{fig:combo1}
\end{figure*}

\subsubsection{End Time and Best Time}

Figure \ref{fig:time_and_epoch} shows average \emph{End Time} and \emph{Best Time} for each model. 
The text-conditioned diffusion models with the shortest training times are 
\texttt{MiniLM\allowbreak-single\allowbreak-regular}, 
\texttt{MiniLM\allowbreak-single\allowbreak-absence}, 
\texttt{MLM\allowbreak-regular}, and 
\texttt{MLM\allowbreak-absence} 
with times from 12.58 to 14.1 hours. 
However, \texttt{MiniLM}'s $\text{c-score}$s on random captions were worse than \texttt{MLM}, so the small time difference is not worth the performance drop. 

In terms of \emph{Best Time}, the same models are fastest, but in different order:
\texttt{MiniLM\allowbreak-single\allowbreak-regular}, 
\texttt{MLM\allowbreak-regular}, 
\texttt{MLM\allowbreak-absence}, and 
\texttt{MiniLM\allowbreak-single\allowbreak-absence}. 
Times range from 11.44 to 12.51 hours.
For these models, the best epoch came slightly before the final epoch. \texttt{GTE\allowbreak-multiple\allowbreak-negative} is the only model where this difference was huge: 126.3 vs.\ 73.97 hours. 

In general, \texttt{negative} captions take more time for little gain. Although some \texttt{GTE} models were comparable to \texttt{MLM} in terms of caption score, they take longer to train, even in terms of \emph{Best Time}. Pretrained sentence transformers that use \texttt{multiple} phrase embeddings take longer to train than their \texttt{single} embedding counterparts.

The models that train the quickest are unconditional diffusion, WGAN, and FDM. However, WGAN and unconditional diffusion offer no text guidance, and FDM performance is much worse, so the extra speed is of little benefit.

\subsubsection{Average Minimum Edit Distance}

To compare the diversity of generated levels, 
four sets of 
scenes are created by most models: scenes from the full set of real game scene captions, 100 samples from this set, scenes from random captions not in the original data, and unconditionally generated scenes. Except for  \emph{real (full)} data, these sets each contain 100 scenes to allow fair comparison. WGAN and unconditional diffusion cannot generate scenes from captions, so they only have unconditional samples. FDM can technically create unconditional samples from an empty embedding vector, but it was not intended to, and the results are so terrible that we do not include them. MarioGPT is a special case, since its scenes were not generated unconditionally, but also were not generated by our captions. We compare against 100 size $16 \times 16$ scenes from MarioGPT as described earlier.

Figure \ref{fig:AMED_self} shows $\text{AMED}_{\text{self}}$ scores. In general, \emph{real (full)} is small because comparing against more scenes makes it more likely to find a similar scene. This is why \emph{real (100)} was needed for fairness, and is always higher. Across text-conditioned diffusion models, \emph{real (full)} is around 6 tiles, and \emph{real (100)} is around 22 tiles, except for \texttt{GTE\allowbreak-multiple\allowbreak-negative} and \texttt{MiniLM\allowbreak-multiple\allowbreak-negative} around 19 tiles. These two models also had poor caption adherence scores.
For reference, \emph{Real data} shows that $\text{AMED}_{\text{self}}$ is 10.4077 tiles 
on the full set of game data 
and 20.87 tiles in the 100 sample case. 
The collection of all real game data shows more diversity
than what models produce using all real game captions, but the diversity across the evenly spaced set of 100 samples is about the same.


\begin{figure*}[t]
\centering

    \begin{subfigure}[t]{0.33\textwidth}
        \centering
        \includegraphics[width=\linewidth]{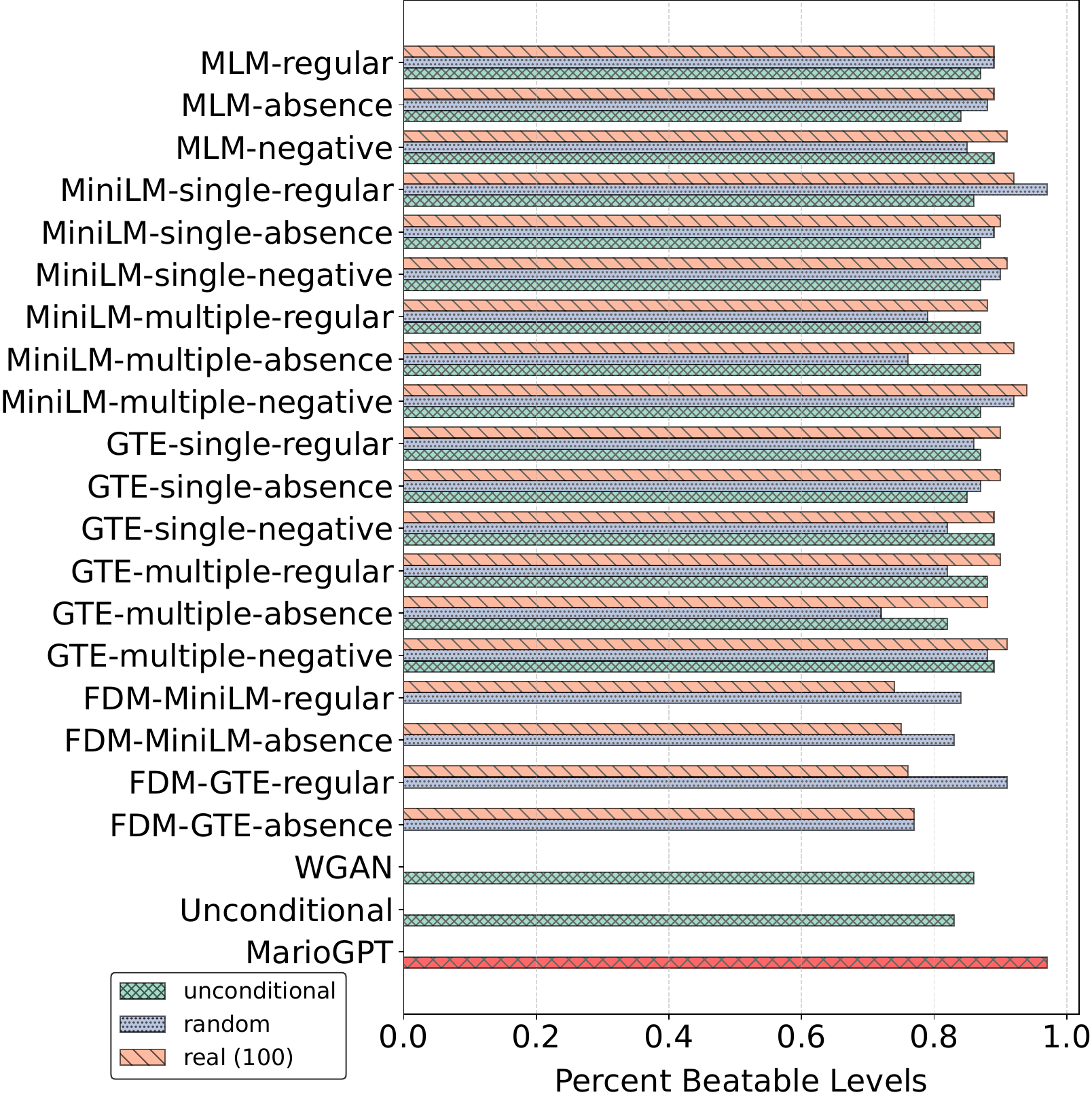} 
        \caption{A* Solvability on Model Outputs}
        \label{fig:astar}
    \end{subfigure}
    \hfill
    \begin{subfigure}[t]{0.33\textwidth}
        \centering
        \includegraphics[width=\linewidth]{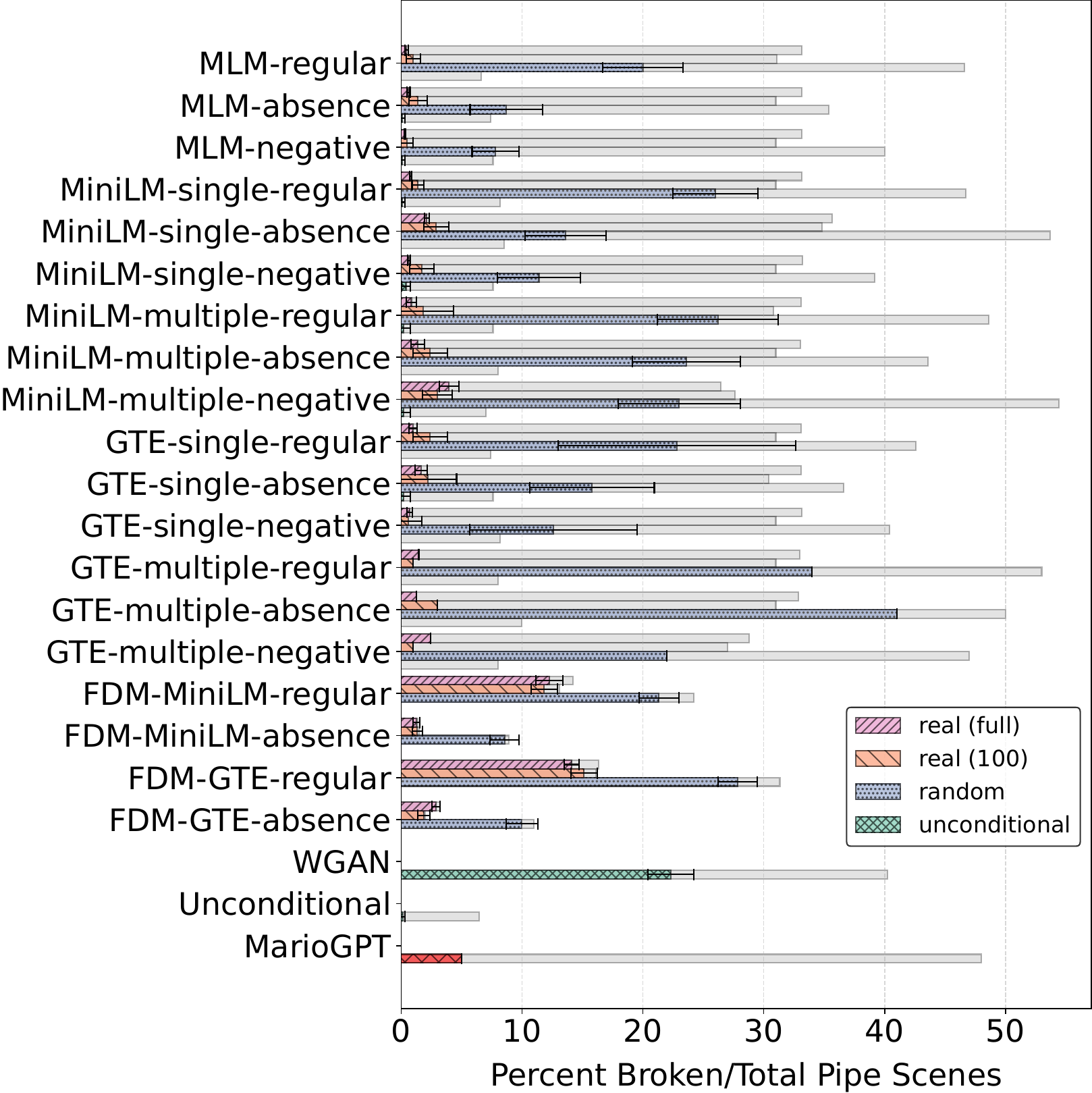} 
        \caption{Percent Scenes With Broken Pipes}
        \label{fig:broken_pipes}
    \end{subfigure}
    \hfill
    \begin{subfigure}[t]{0.33\textwidth}
        \centering
        \includegraphics[width=\linewidth]{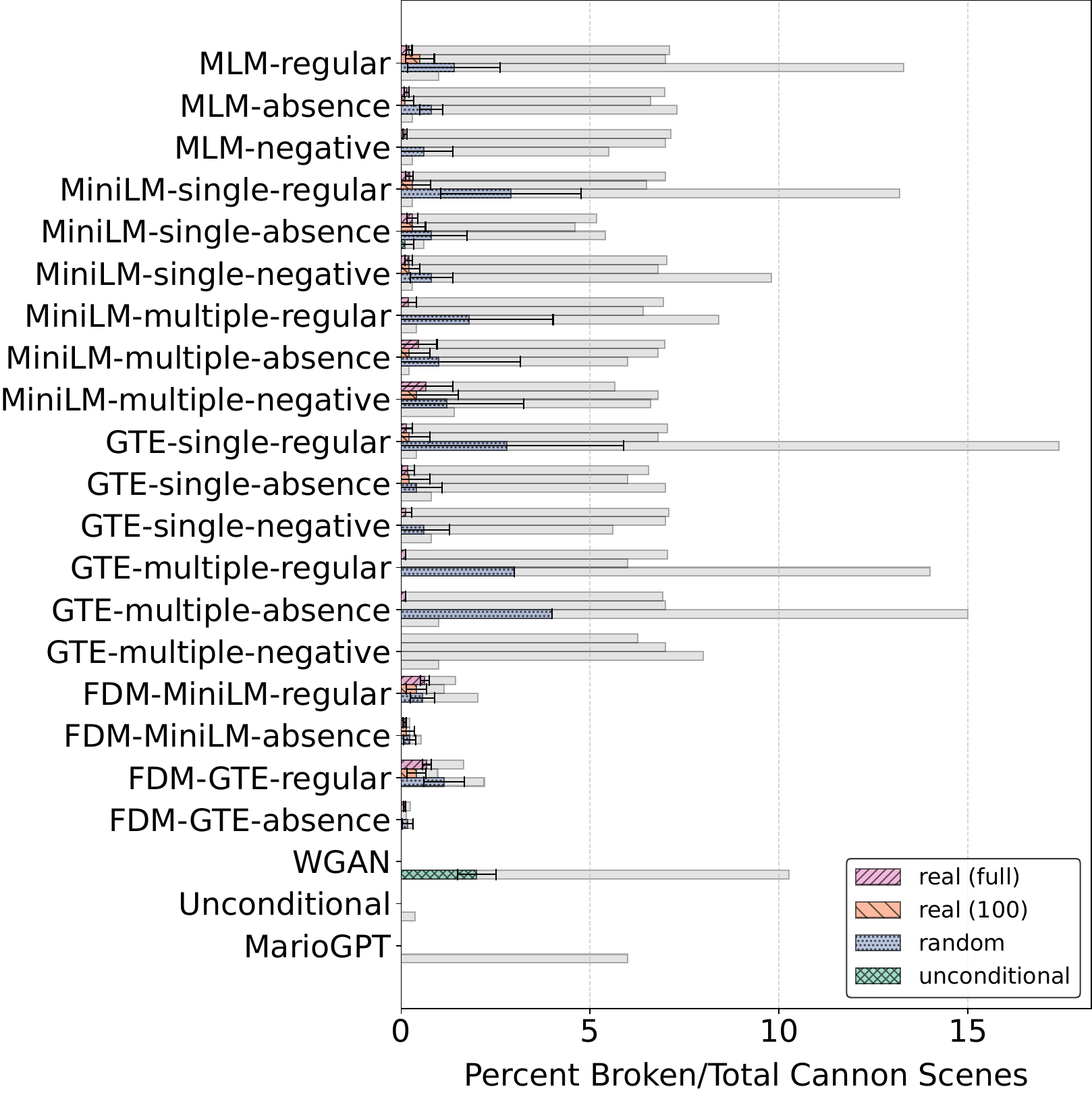} 
        \caption{Percent Scenes With Broken Cannons}
        \label{fig:broken_cannons}
    \end{subfigure}

\caption{
(\subref{fig:astar}) A* results show the percentage of beatable  scenes. Only one model and 100 scenes of each type are used, so there are no error bars.
(\subref{fig:broken_pipes}) For each dataset, colored bars with 95\% confidence intervals show the percentage of scenes that contain a broken pipe. The gray bar extending from each bar represents the scenes with pipes of any kind: regular, broken, or upside down. Producing few broken pipes is more impressive if valid pipes are also generated, as indicated by larger gray regions. Most models produce few broken pipes, but more with random captions. \texttt{FDM\allowbreak-regular} models produce more broken pipes than \texttt{FDM\allowbreak-absence}. WGAN produces many broken pipes.
(\subref{fig:random_negative}) Broken cannon data is presented similarly. Many datasets have no broken cannons, though cannons in general are rarer than pipes.
}
\label{fig:combo2}
\end{figure*}

For \emph{random} captions, diversity varies more but is generally higher than for real captions, though there are exceptions: \texttt{MLM\allowbreak-absence} and \texttt{MLM\allowbreak-negative}. Some models with lower caption adherence have higher diversity from random captions; poor level structure can result in high edit distances, since tiles will be in weird places. Unconditional samples are generally less diverse than caption-generated samples, the only exception being \texttt{GTE\allowbreak-multiple\allowbreak-negative}. WGAN and unconditional diffusion have comparable scores to the unconditional samples from text-conditioned diffusion models. FDM's $\text{AMED}_{\text{self}}$ scores are extremely low in all categories. MarioGPT's score is around 25, which is higher than \emph{real (100)} for all models but lower than the \emph{random} score of most.

Figure \ref{fig:AMED_real} compares $\text{AMED}_{\text{real}}$ results. Samples from random captions have high distances, but distances from real captions are much smaller, with unconditional samples usually between. \emph{real (full)} and \emph{real (100)} are generally closer to each other, around 3-5 tiles, though some FDM results are higher. In other words, for captions seen during training, models often create nearly identical scenes, which is probably for the best, but it means that alternate scenes that share the caption are not generated. This is a limitation in how well models generalize. Thankfully, high \emph{random} scores indicate that generation of novel scenes is possible, though being too novel caries the risk of being disorganized, as occurs with WGAN results. WGAN samples have higher $\text{AMED}_{\text{real}}$ scores because they struggle to fit the data. In contrast, samples from unconditional diffusion have $\text{AMED}_{\text{real}}$ scores similar to unconditional samples from text-conditioned diffusion models. FDM is once again an anomaly, since its \emph{ramdom} scores are much lower. 
MarioGPT's scores are higher than most, though the highest \emph{random} scores of certain diffusion models are more novel.

\subsubsection{Solvability}

Figure \ref{fig:astar} shows the percentage of beatable scenes from each model. Since simulation takes time, we only apply A* to 100 samples from real captions per one model of each type as opposed to all model outputs across all real captions. Even the worst, \texttt{GTE\allowbreak-multiple\allowbreak-absence} with random captions, produces 72\% beatable scenes. FDM results from real captions are between 74-77\%. The highest score is a tie at 97\% between MarioGPT and \texttt{MiniLM\allowbreak-single\allowbreak-regular}'s random caption samples. Most diffusion models have scores between 82\% and 94\%.

\subsubsection{Level Integrity}

Figure \ref{fig:broken_pipes} shows the percentage of generated scenes that contain one or more broken pipes and the percentage of scenes with any kind of pipe. 
For datasets with 100 samples, the percentage is also the count.
Most models produce few scenes with broken pipes using real captions, but many using random captions. Captions produce many \emph{valid} pipes too. FDM creates more broken pipes with \texttt{regular} captions than \texttt{absence} captions.
MarioGPT has 5 broken pipe scenes, which is more than
diffusion models on real captions, though less than diffusion models on random captions.
WGAN produces many broken pipes, but unconditional scenes from diffusion models have almost no broken pipes, though they have fewer valid pipes as well.

Figure \ref{fig:broken_cannons} shows similar results for broken cannons. Cannons are rarer, and broken ones rarer still, though more broken cannons are associated with random captions and WGAN. Unconditional samples have almost no broken cannons, but very few valid ones either.


\section{Larger Levels}


Although the diffusion models are trained on $16 \times 16$ samples, the output size can be any value during inference, making it possible to generate levels of arbitrary length. However, there are two problems. First, input prompts are calibrated for smaller scenes, so it is not clear what one should expect from larger scenes, or how to assess them. Secondly, longer levels seem to often be unbeatable due to levels having massive gaps that cannot be jumped over.

\begin{figure}[t]
\centering
\includegraphics[width=\linewidth]{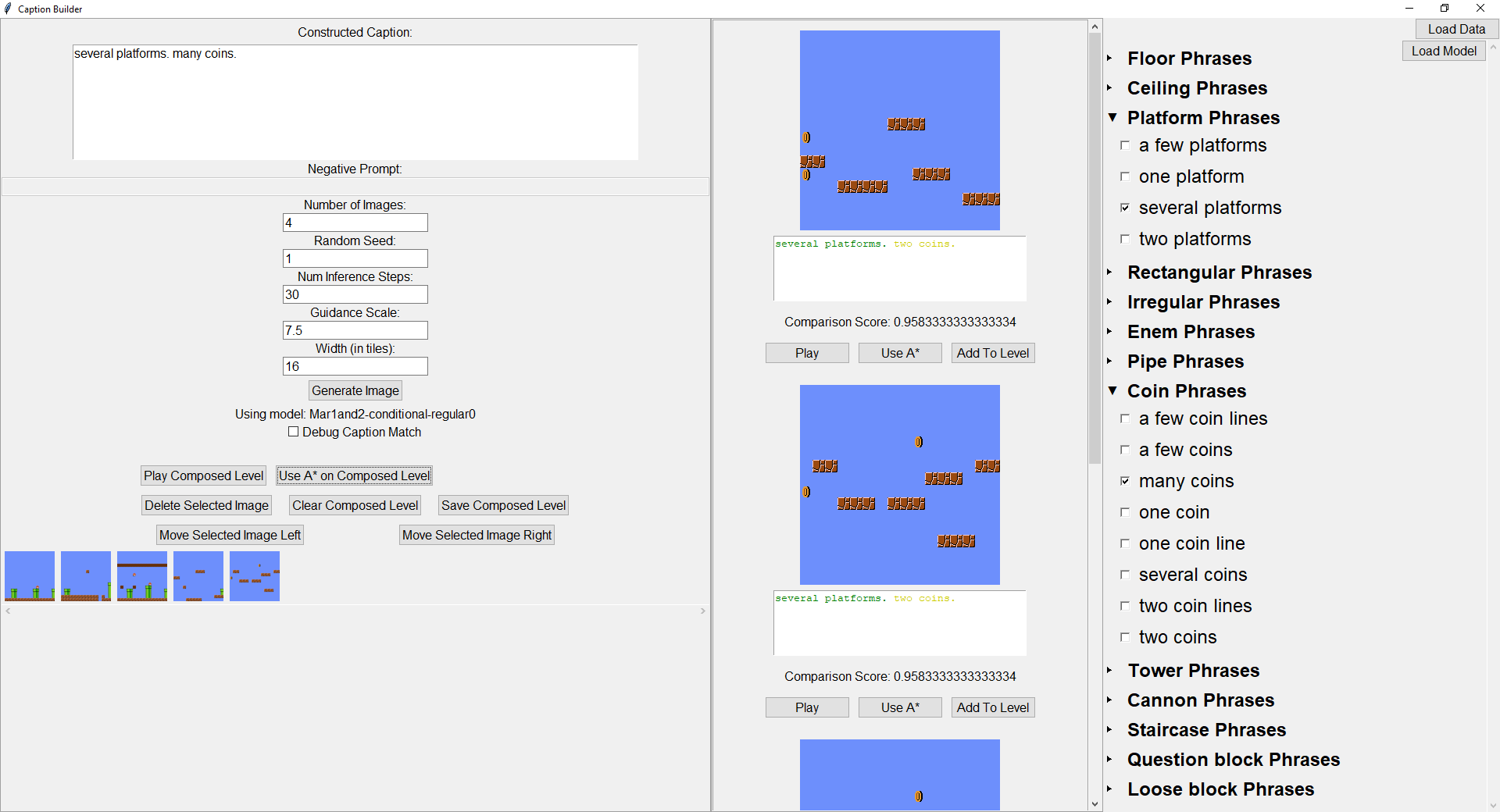} 
\caption{Interactive Level Building Interface. Phrases are selected from the right which create a caption in the upper-left. Then samples can be generated which appear in the center column. These can be added to a level in the lower-left which can be played by a human or A* agent.}
\label{fig:gui}
\end{figure}

It may be possible to address these issues with a dataset consisting of longer levels, but confirming this is a task for future work. In the meantime, human designers can still benefit from tools we have designed by incorporating diffusion models into a mixed-initiative system with a GUI for building complete levels from diffusion-generated scenes.
The interface (Figure \ref{fig:gui}) supplies checkboxes for valid phrases organized by topic (floor, coins, etc.) so that creating descriptive captions is easy to do without worrying about spelling or the use of unknown vocabulary. Once a caption is constructed, level scenes can be generated with a chosen model.
Parameters like the random seed, number of samples, number of inference steps, guidance scale, and scene width can be set. Generated scenes have automatically generated captions that are color-coded to visually indicate differences/similarities to the user-supplied prompt. Caption adherence score is also displayed. 
Individual scenes can be combined into a larger level. Scenes are added to the end of the level by default, but they can be moved or deleted by the user. 
Both constructed levels and individual scenes
can be tested via human play or with A*. Finally, ASCII text versions of constructed levels can be saved for future use. Examples of longer composed levels are in the online appendix, as is a video demonstrating how to use the GUI: \url{https://people.southwestern.edu/~schrum2/mario.html}

Co-creative level design systems have been explored in the academic literature before \cite{larsson:aiide2022},
including for Mario \cite{guzdial:aiideEXAG2018, schrum2020interactive}.
Evolutionary computation and various types of AI models are common in such work, but the ability to mix and match components produced from text prompts is new, and opens up possibilities for numerous human subject studies. Experimental verification of the benefits of such a system is still needed, but the authors found it easy to create scenes of different sizes and combine them into fun and interesting levels of arbitrary length.
The generation of multiple scenes per caption along with the ability to change the random seed and guidance scale make it easier to find scenes that match a desired caption, and users may also be serendipitously inspired by scenes they create, even when they do not match the input caption. We hope to study the experiences of users interacting with this system in future work.


\section{Limitations}


Our approach to diffusion-based PCGML requires a sufficiently sized dataset of level scenes and enough expert knowledge and programming skill to assign adequate captions to such scenes algorithmically. However, we believe this is a modest barrier and are actively expanding our research to other games. 
Although we used NVIDIA GPUs, they were not excessively expensive; a decent gaming PC is able to train our models.
Long levels are less likely to be beatable, but our GUI provides an effective way to make beatable long levels with slight additional effort.

\section{Discussion and Conclusion}

It was surprising that a small and basic transformer architecture with little training on a small dataset with limited vocabulary produced the best results when combined with our diffusion model. It was also disappointing that attempts to enhance the approach had little effect or a detrimental effect. Models take longer to train with \texttt{negative} prompts, and are not better. The \texttt{absence} captions are more complicated, but offer no benefit. \texttt{MiniLM} and \texttt{GTE} are more powerful and general language models, but do not produce definitively better results, which is especially damning in the case of \texttt{GTE}, whose models take much longer to train. We thought \texttt{multiple} sentence embeddings could provide richer context for diffusion, but they simply increased training time.
Of course, it is not bad that a simpler model can be so effective, though we wonder if other pretrained language models or alternative architectures might be better.

There must be some way to break the average caption score barrier of 0.5 on random captions.
In fairness, some of the random captions are very unusual. 
Actual captions in the dataset include ``two ascending staircases.'' and ``two question blocks. two enemies. two cannons.'' These captions do not mention a floor. Although levels without floors exist, such levels tend to have many platforms to support entities like cannons and enemies.

We tested larger UNet architectures in our preliminary experiments, but they only seemed to increase training time with no clear benefit. We admit that a more systematic exploration of different architectures and hyperparameters could lead to improvements. However, we suspect the biggest limitation is the training data, but part of that problem could be solved with our automatic captioning system. Although our model may not always produce scenes with desired captions, it can produce scenes with captions that do not exist in the dataset. If such scenes were to be automatically captioned and added to the training dataset, it may be possible to gradually accumulate enough scenes to get near complete coverage of the space of captions. Our results suggest that once a caption is in the dataset, our model would have no trouble generating a scene that matches it.

MarioGPT does well in several metrics, though our best diffusion models are comparable or superior in certain instances. However, it is unclear what the best comparison is: real, random, or unconditional. As stated above, our diffusion models do well with captions they've seen during training; MarioGPT's limited caption options makes it less likely to see an unfamiliar caption.
It would be interesting to compare diffusion with simpler captions, or MarioGPT with more complex captions.

For now, we conclude, having presented a method to make captions for Mario level scenes which allows for the training of a transformer text encoder and diffusion model that can make realistic Mario scenes, which can be combined into larger levels using our GUI.

\section{Acknowledgments}

This paper was initially drafted without AI, but ChatGPT was later used to improve clarity/concision. We thank donors to Southwestern University's SURF program for support.


\clearpage

\appendix
\section{Appendix}

\begin{figure*}[!t] 
  \centering
  \includegraphics[width=\textwidth]{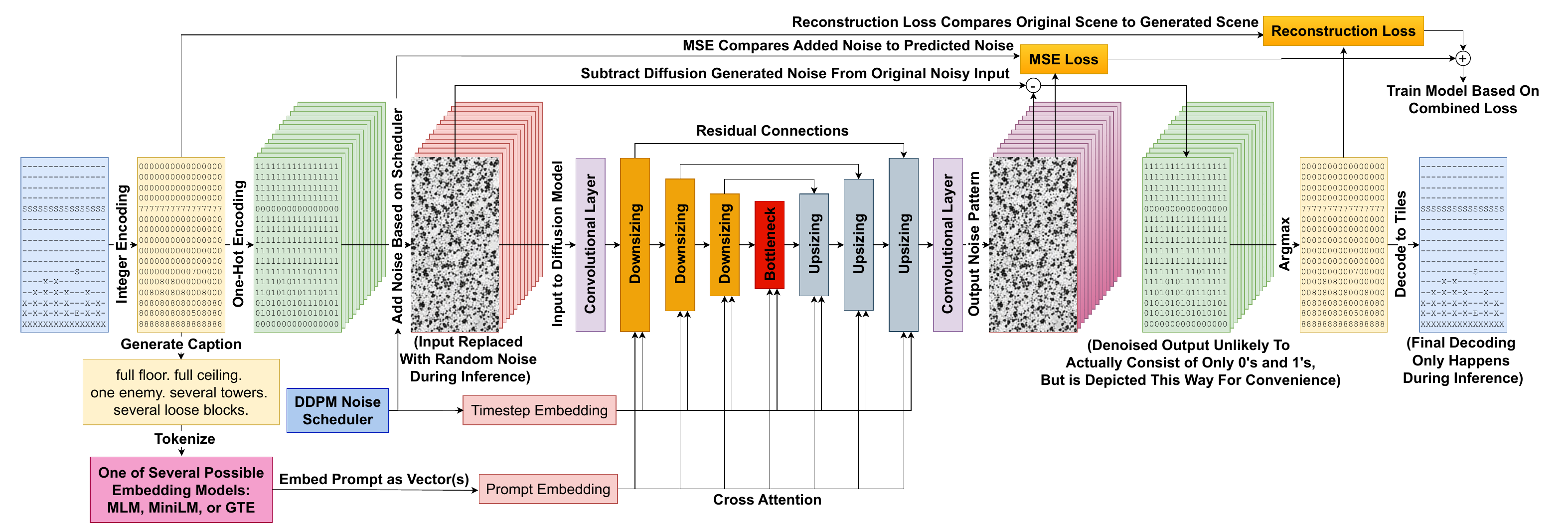}
  \caption{Diffusion Training and Inference Pipeline. Our training set is integer-encoded, but was derived from the ASCII data in the VGLC. Each scene is associated with an automatically generated caption. The scenes are one-hot encoded before noise is added according to a DDPM scheduler. The noisy input enters the diffusion model, while its cross-attention blocks access a hidden state based on
  both a timestep embedding from DDPM and a prompt embedding from whichever text model is being used. The output of the model is a noise prediction which is directly compared to the known amount of added noise to complete the Mean Squared Error. The predicted noise is also removed from the noisy input to approximate the one-hot
  encoded training data, which is then integer-encoded via argmax
  for comparison to the original training sample. This is how the reconstruction loss is calculated. Both losses are combined to train the model.}
  \label{fig:pipeline}
\end{figure*}

Additional hyperparameter settings and results that are only in the arXiv pre-print.

\subsection{Dataset Details}

Our cleaned version of the VGLC and our captioning approach resulted in data with the following properties:

\begin{itemize}
    \item Number of Super Mario Bros.\ levels: 20
    \item Number of Super Mario Bros.\ 2 levels: 22
    \item Total $16 \times 16$ samples across both games: 7,687
    \item Vocabulary size for \texttt{regular} captions: 47
    \item Vocabulary size for \texttt{absence} captions: 48
    \item Training samples: 6918
    \item Validation samples: 384
    \item Test samples: 385
\end{itemize}

The tiles available in Mario levels are in Table~\ref{tab:tiles}.

\subsection{Text Encoder Details}

These details are relevant to our \texttt{MLM} model:

\begin{itemize}
    \item Token embedding size: 128
    \item Number of transformer encoder layers: 4
    \item Number of attention heads: 8
    \item Dimension of hidden layer: 256
    \item Probability of [MASK] token during MLM training: 0.15
    \item Training optimizer: AdamW
    \item Training epochs: 300
    \item Loss function: Cross Entropy Loss
    \item Learning rate: Starts at 0.00005
    \item Minimum learning rate: 0.000001
    \item Learning rate schedule: ReduceLROnPlateau
    \item Training batch size 16
\end{itemize}

\begin{table}[t]
\centering
\caption{\label{tab:tiles}Tile types in Mario levels.  
Symbol characters come from the VGLC. 
Identity values are used for one-hot encoding.
Visualizations are used by the
Mario AI framework.}
\begin{tabular}{cccc}
\hline
Tile type & Symbol & Identity & Visualization\\
\hline 
Empty/Sky (passable) & - & 0 & \includegraphics[scale=0.5]{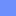} \\
Top-left pipe & $<$ & 1 & \includegraphics[scale=0.5]{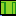}\\
Top-right pipe & $>$ & 2 & \includegraphics[scale=0.5]{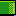}\\
Full question block & ? & 3 & \includegraphics[scale=0.5]{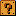}\\
Cannon top & B & 4 & \includegraphics[scale=0.5]{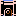}\\
Enemy & E & 5 & \includegraphics[scale=0.5]{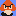}\\
Empty question block & Q & 6 & \includegraphics[scale=0.5]{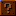}\\
Breakable & S & 7 & \includegraphics[scale=0.5]{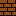}\\
Solid/Ground & X & 8 & \includegraphics[scale=0.5]{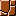}\\
Left pipe & [ & 9 & \includegraphics[scale=0.5]{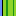}\\
Right pipe & ] & 10 & \includegraphics[scale=0.5]{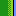}\\
Cannon support & b & 11 & \includegraphics[scale=0.5]{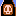}\\
Coin & o & 12 & \includegraphics[scale=0.5]{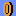}\\
\hline
\end{tabular}
\end{table}

\begin{figure*}[t]
\centering

    \begin{subfigure}[t]{0.33\textwidth}
        \centering
        \includegraphics[width=\linewidth]{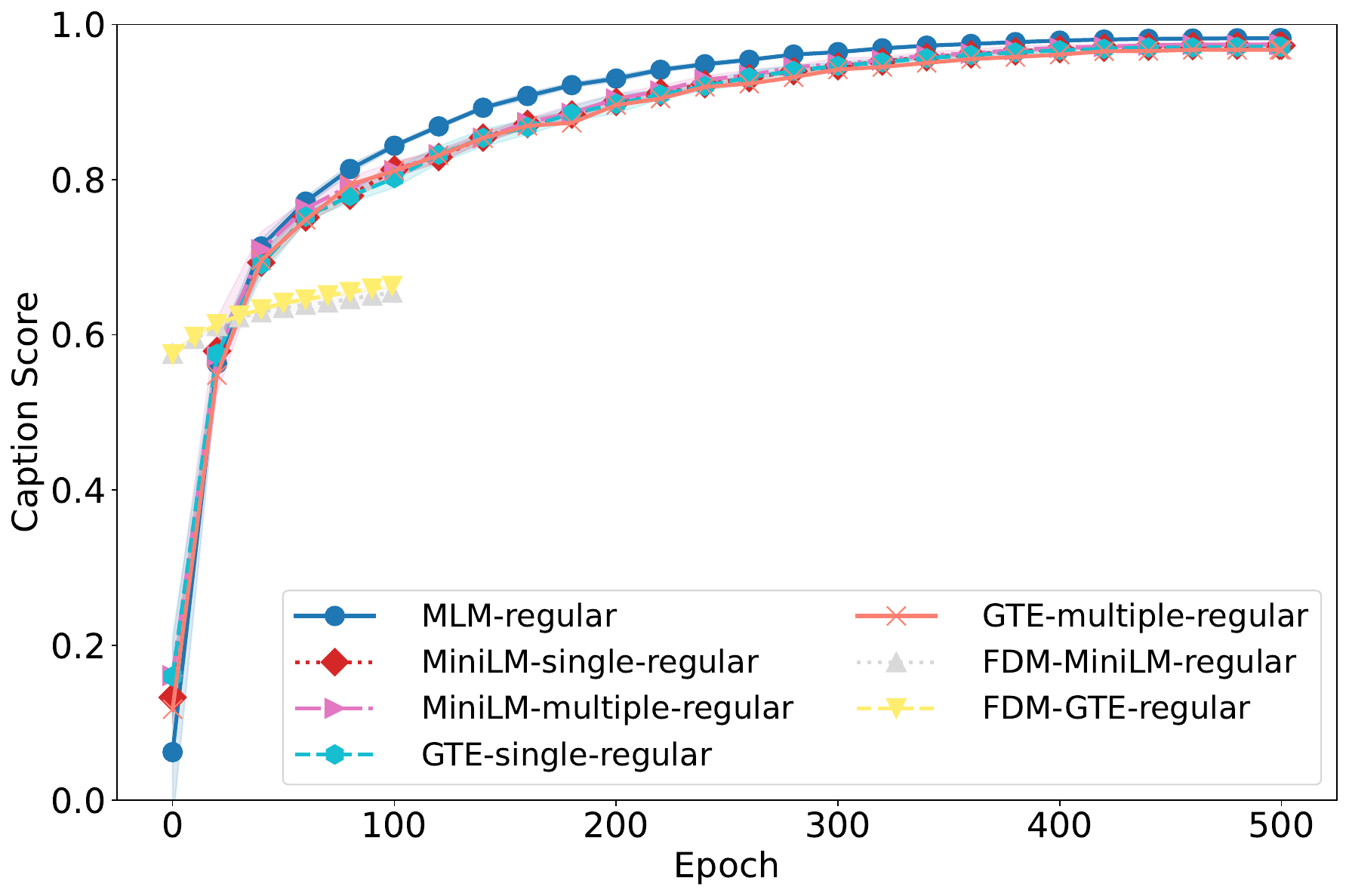} 
        \caption{\texttt{regular}}
        \label{fig:realfull_regular}
    \end{subfigure}%
    \hfill
    \begin{subfigure}[t]{0.33\textwidth}
        \centering
        \includegraphics[width=\linewidth]{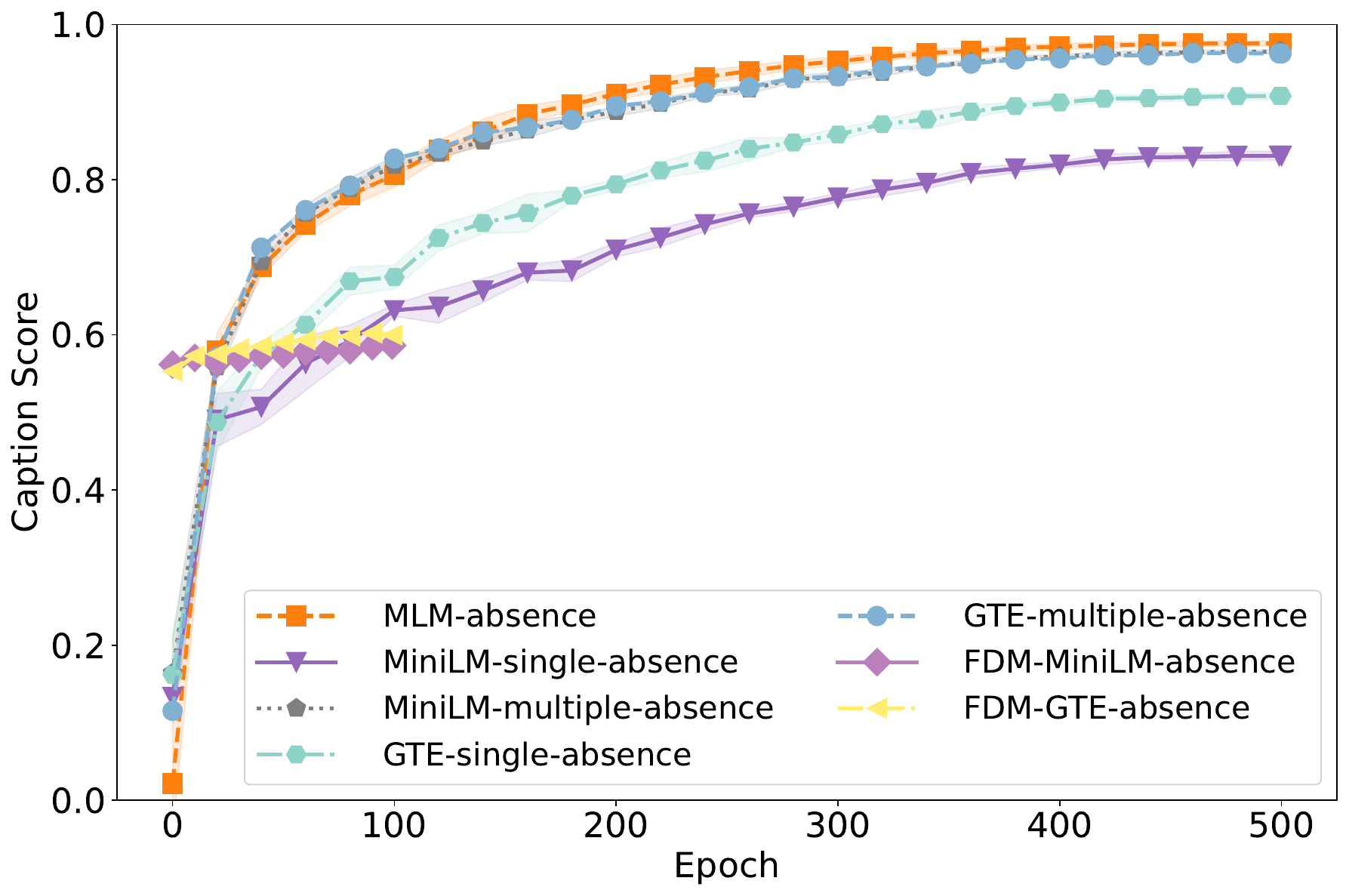} 
        \caption{\texttt{absence}}
        \label{fig:realfull_absence}
    \end{subfigure}
    \hfill
    \begin{subfigure}[t]{0.33\textwidth}
        \centering
        \includegraphics[width=\linewidth]{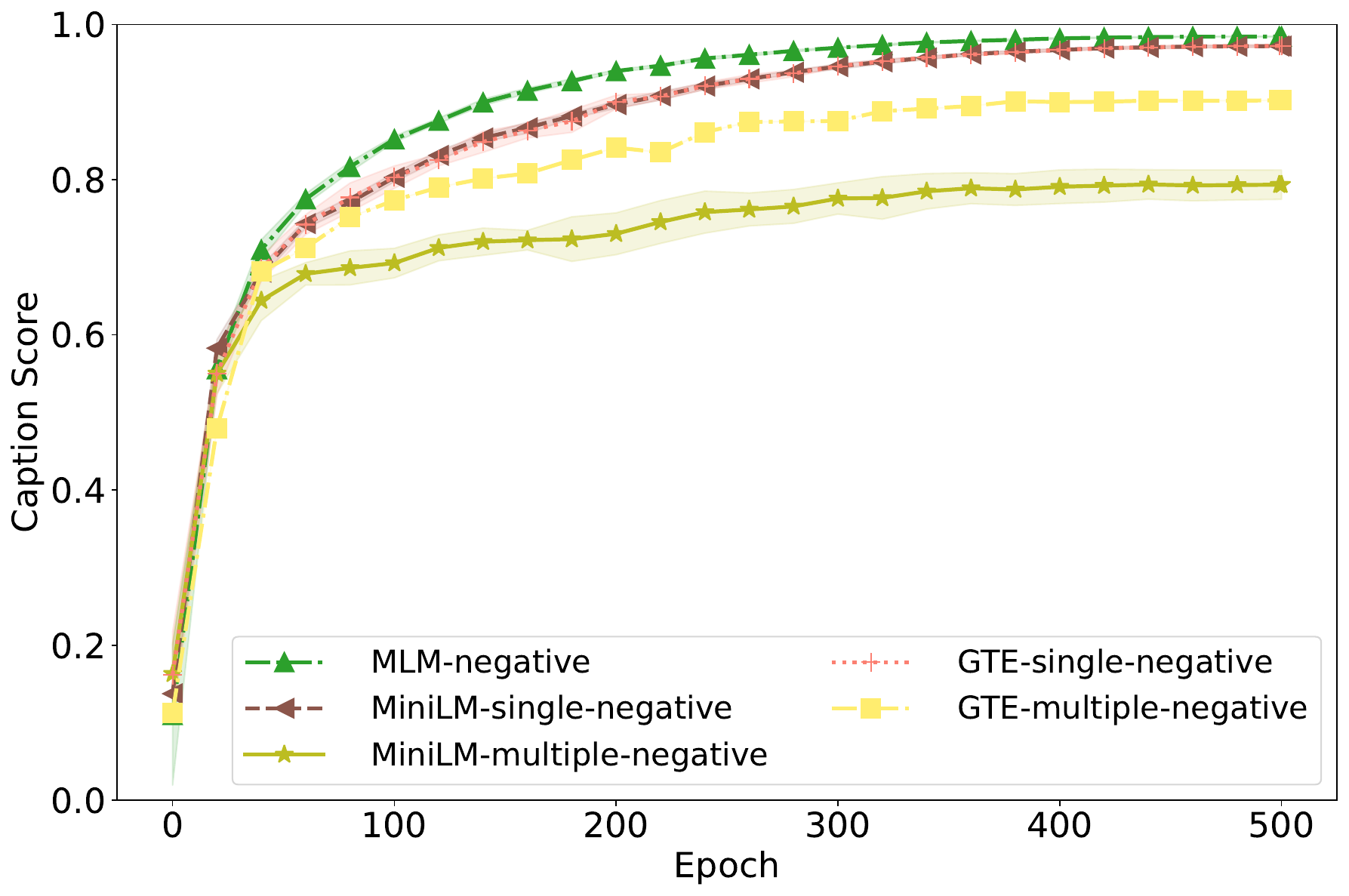} 
        \caption{\texttt{negative}}
        \label{fig:realfull_negative}
    \end{subfigure}

\caption{Average Caption Adherence Score by Epoch on All Real Game Captions. 
Results are qualitatively similar to those in Figure \ref{fig:caption_real}.
(\subref{fig:realfull_regular}) \texttt{regular} caption results.
(\subref{fig:realfull_absence}) \texttt{absence} caption results.
(\subref{fig:realfull_negative}) \texttt{negative} caption results.}
\label{fig:caption_real_full}
\end{figure*}

\subsection{Diffusion Model Details}

These details are relevant to our diffusion models:

\begin{itemize}
    \item Base dimension of the UNet: 128
    \item Number of residual blocks for downsampling: 2
    \item UNet encoder (down) channels: 13, 128, 256, 512
    \item UNet decoder (up) channels: 512, 256, 128, 13
    \item Number of attention heads: 8
    \item Noise schedule: DDPM with a linear beta schedule
    \item Noise betas: 0.0001 to 0.02
    \item Noise schedule time steps: up to 1000
    \item Training optimizer: AdamW 
    \item AdamW weight decay: 0.01 
    \item AdamW beta values: 0.9 and 0.999
    \item Gradient accumulation steps: 1
    \item Learning rate schedule: cosine
    \item Learning rate warm-up period: 25 epochs
    \item Top learning rate: 0.0001
    \item Guidance scale during inference: 7.5
    \item Inference steps: 30
\end{itemize}

The loss function for the diffusion model is the same one used by \citeauthor{lee:mva23} (\citeyear{lee:mva23}):

\begin{align}
\mathcal{L}_{\text{total}} &= \mathcal{L}_{\text{MSE}} + \lambda \mathcal{L}_{\text{rec}} \\
\mathcal{L}_{\text{MSE}} &= \frac{1}{N} \sum_{i=1}^N \| \hat{\epsilon}_i - \epsilon_i \|^2 \\
\mathcal{L}_{\text{rec}} &= -\frac{1}{N} \sum_{i=1}^N \sum_{h=1}^H \sum_{w=1}^W \log P_{\theta}(O_{i,h,w} | x_{i,h,w})
\end{align}

\noindent where $\lambda = 0.001$ is the weight on the reconstruction loss,
$N$ is the batch size,
$\hat{\epsilon}_i$ is the model's predicted noise for sample $i$,
$\epsilon_i$ is the true noise,
$H$ and $W$ are the height and width of 16,
$O_{i,h,w}$ is the ground truth for the tile at position $(h, w)$ in sample $i$,
$x_{i,h,w}$ is the generated tile at position $(h, w)$ in sample $i$,
so $P_{\theta}(O_{i,h,w} | x_{i,h,w})$ is the probability of the original
block given the generated block according to the diffusion model with parameters $\theta$.

Figure \ref{fig:pipeline} depicts the complete diffusion pipeline for training and inference.

\subsection{Five-Dollar Model Details}

These details are relevant to our Five-Dollar Models:

\begin{itemize}
    \item Number of residual blocks: 3
    \item Number of convolutional filters: 128
    \item Kernel size: 7, but 3 for final layer
    \item Noise vector size: 5 
    \item Training epochs: 100
    \item Loss function: Negative Log Likelihood Loss
    \item Training optimizer: AdamW
    \item Learning rate: 0.001
\end{itemize}

\subsection{Additional Performance Metrics and Results}

Results dealing with these performance metrics could not fit into the main text of the paper.

\subsubsection{Caption Adherence on Full Dataset}

When applied to the set of all captions from the original games (Figure \ref{fig:caption_real_full}), the caption adherence score is qualitatively similar to the results from just the test set data, as demonstrated earlier (Figure \ref{fig:caption_real}).

\subsubsection{End Time and Best Time on Logarithmic Scale}

\begin{figure}[t]
\centering
\includegraphics[width=0.9\linewidth]{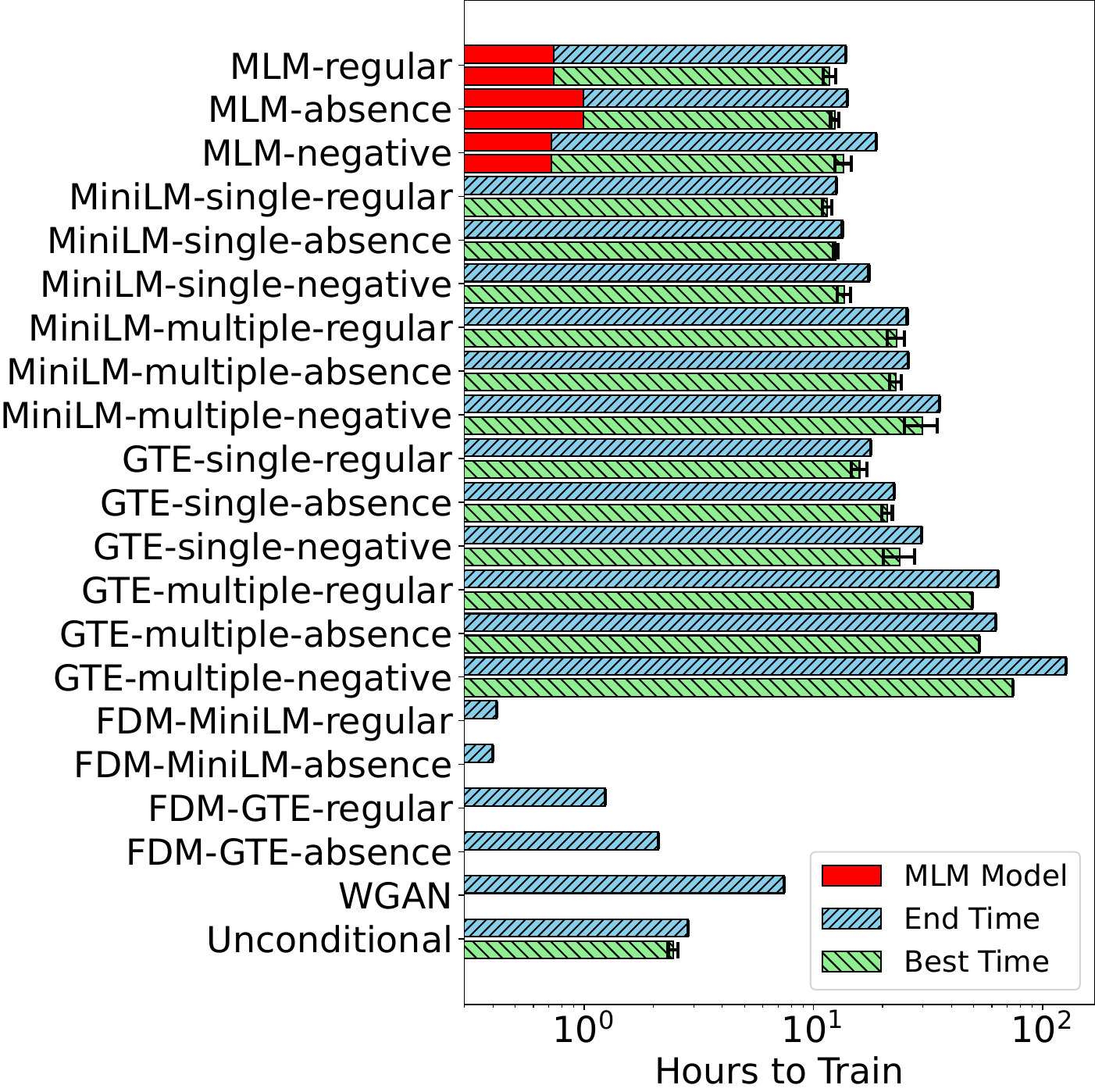} 
\caption{Average End Times and Best Times on Log Scale.}
\label{fig:timing_log}
\end{figure}

Most execution times are small, but a few larger values skew the presentation in Figure \ref{fig:time_and_epoch}. The same data from that figure is depicted in Figure \ref{fig:timing_log} using a logarithmic scale.

\subsubsection{Caption Order Tolerance}

\begin{figure}[t]
\centering
\includegraphics[width=0.9\linewidth]{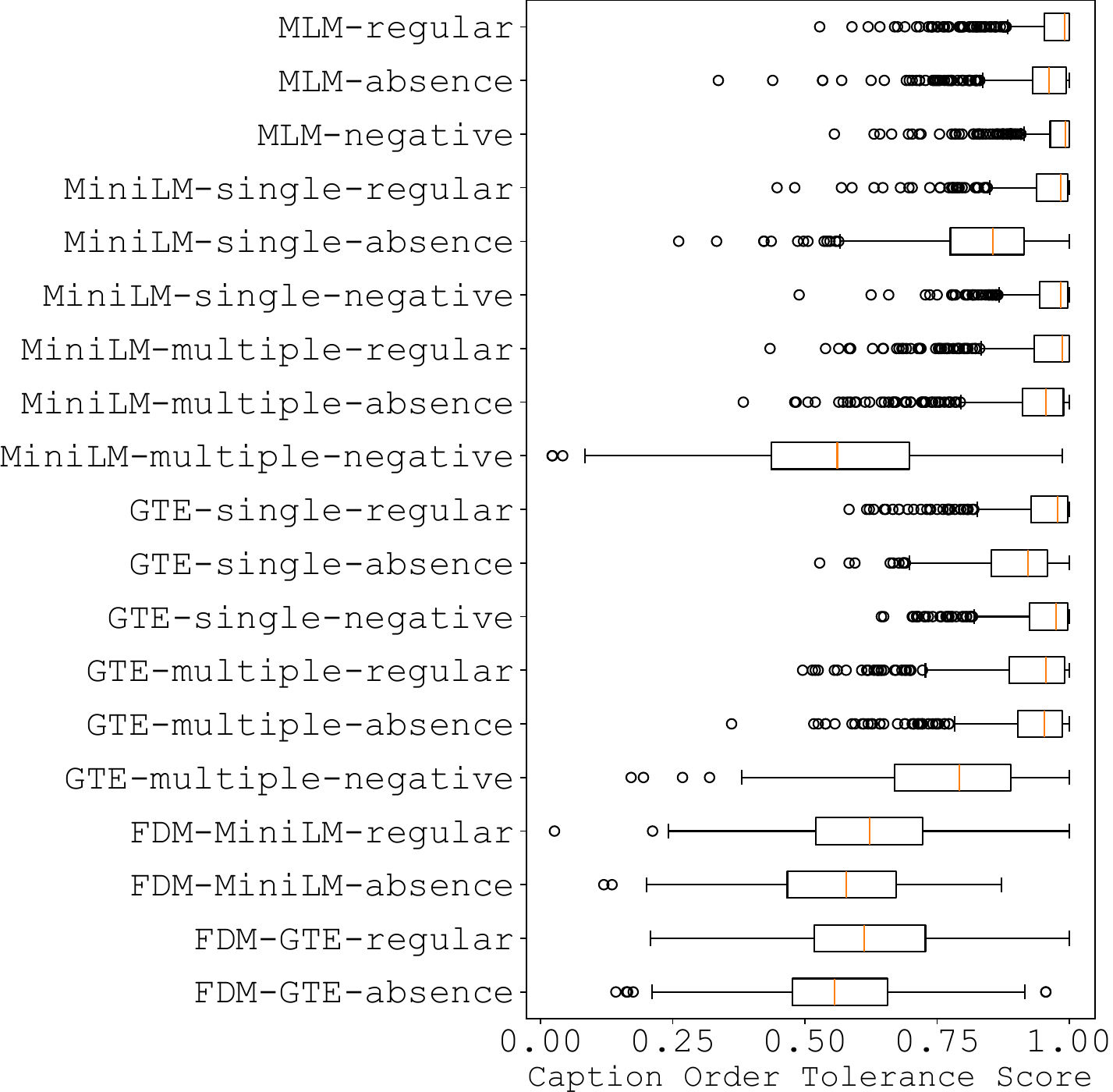} 
\caption{Caption Order Tolerance.
Shows how models handle different phrase orderings in captions of real game scenes. One model of each type is considered with scores for each caption in the test set. Most models do well, except \texttt{MiniLM\allowbreak-single\allowbreak-absence}, \texttt{MiniLM\allowbreak-multiple\allowbreak-negative}, \texttt{GTE\allowbreak-multiple\allowbreak-negative}, and FDM.}
\label{fig:caption_order_tolerance}
\end{figure}

We want to give users the flexibility to provide caption phrases in whatever order they prefer. Semantically, a caption is equivalent to any caption that is a permutation of its phrases. We can take a caption and sample some number of its permutations, send each one through a text-to-level model, and average the $\text{c-score}$s: 

\begin{equation}
\label{ref:tolerance}
\text{tolerance}(P) = \frac{\sum_{(p,c) \in P} \text{c-score}(p,c)}{|P|} 
\end{equation}

\noindent $P$ is a set of pairs $(p,c)$, where $p$ is a prompt and $c$ is the caption on the level a model produces using $p$. Values of $p$ are distinct permutations of the same input prompt. 

Prompts can contain many phrases, so averaging across all permutations would be computationally expensive. Instead, we sample up to 5 distinct random permutations per prompt.

Caption order tolerance results are in Figure \ref{fig:caption_order_tolerance}.

\subsection{Making Larger Levels}

Tables \ref{tab:long_level}, \ref{tab:long_segment_level}, and \ref{tab:long_long_segment_level} show different examples of using the interactive GUI to create longer levels.









\begin{table*}[h!]
\centering
\caption{Long Level Generated One Scene At a Time (16 wide). 
Using \texttt{MLM-regular0} (\url{https://huggingface.co/schrum2/MarioDiffusion-MLM-regular0}), 
the GUI was used to generate $16 \times 16$ scenes with the designated prompts. The segments
were then combined into a single playable level. The caption adherence score
of each scene is shown beneath it.}
\label{tab:long_level}
\setlength{\tabcolsep}{0pt} 
\begin{tabular}{|*{8}{>{\centering\arraybackslash}p{0.125\textwidth}|}}
\hline
full floor. one platform. two enemies. one pipe. a few coins. & 
floor with one gap. a few enemies. a few pipes. one tower. &
floor with several gaps. two platforms. one rectangular block cluster. a few enemies. many coins. one tower. one ascending staircase. &
floor with several gaps. one rectangular block cluster. one irregular block cluster. several enemies. many coins. one tower. one ascending staircase. one descending staircase. two question blocks. &
a few platforms. several enemies. one question block. two loose blocks. &
giant gap with one chunk of floor. a few platforms. one rectangular block cluster. two enemies. one pipe. a few coins. a few loose blocks. &
giant gap with one chunk of floor. a few platforms. a few enemies. a few coins. one coin line. one question block. a few loose blocks. &
floor with one gap. one ascending staircase. \\
\hline
\includegraphics[width=0.125\textwidth]{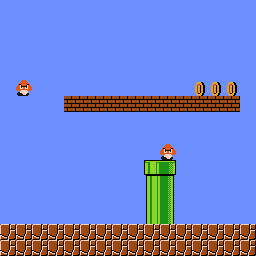} & 
\includegraphics[width=0.125\textwidth]{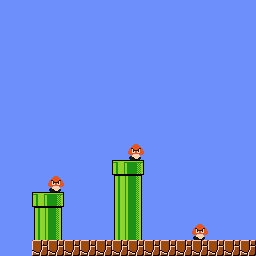} & 
\includegraphics[width=0.125\textwidth]{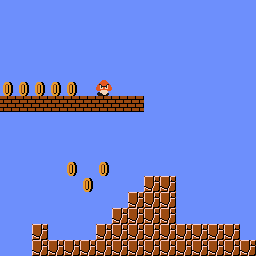} & 
\includegraphics[width=0.125\textwidth]{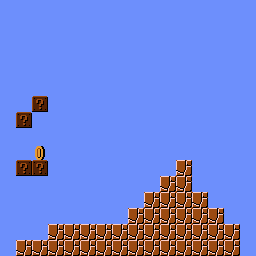} & 
\includegraphics[width=0.125\textwidth]{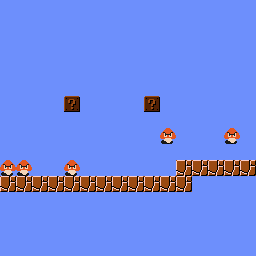} & 
\includegraphics[width=0.125\textwidth]{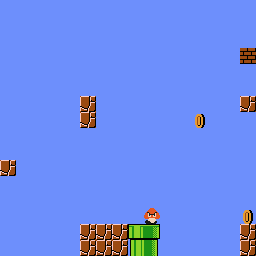} & 
\includegraphics[width=0.125\textwidth]{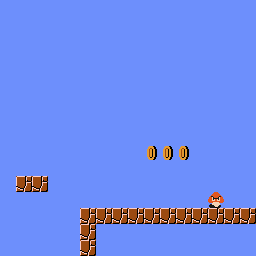} & 
\includegraphics[width=0.125\textwidth]{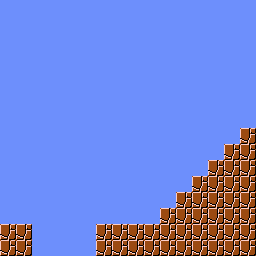} \\
\hline
0.888889 &
0.875 &
0.347222 &
0.458333 &
0.972222 &
0.722222 &
0.819444 &
1.0 \\
\hline
\end{tabular}
\end{table*}

\begin{table*}[ht]
\centering
\caption{Long Level Generated One Scene At a Time (32 wide). 
Using \texttt{MLM-regular0} (\url{https://huggingface.co/schrum2/MarioDiffusion-MLM-regular0}), 
the GUI was used to generate $32 \times 16$ scenes with the designated prompts. The segments
were then combined into a single playable level. Each segment of width 32 has its own
caption adherence score, but the result of splitting each segment into two scenes of
width 16 and averaging those caption adherence scores is also shown. It is generally harder to control the output and to interpret the meaning of the caption adherence scores when the width increases.}
\label{tab:long_segment_level}
\setlength{\tabcolsep}{0pt} 
\begin{tabular}{|*{4}{>{\centering\arraybackslash}p{0.25\textwidth}|}}
\hline
full floor. full ceiling. one enemy. many coins. one coin line. several towers. &
floor with two gaps. ceiling with two gaps. one rectangular block cluster. one irregular block cluster. a few enemies. two pipes. many coins. &
giant gap with several chunks of floor. ceiling with one gap. two irregular block clusters. one enemy. one upside down pipe. many coins. &
floor with two gaps. ceiling with one gap. one platform. two cannons. a few question blocks. \\
\hline
\includegraphics[width=0.25\textwidth]{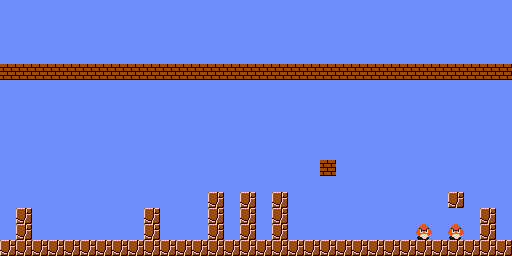} & 
\includegraphics[width=0.25\textwidth]{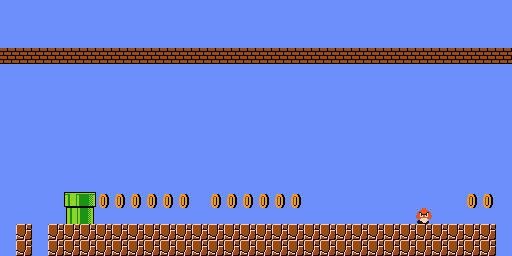} & 
\includegraphics[width=0.25\textwidth]{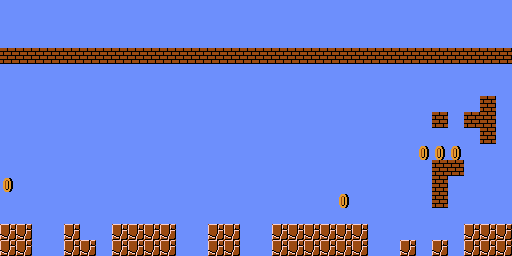} & 
\includegraphics[width=0.25\textwidth]{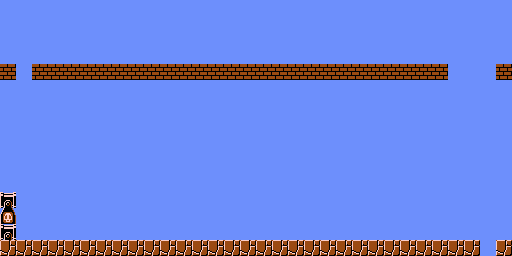} \\
\hline
0.63888889 &
0.56111111 &
0.38055556 &
0.75 \\
AVG: 0.56944444 &
AVG: 0.46388889 &
AVG: 0.33888889 &
AVG: 0.69027778 \\
\hline
\end{tabular}
\end{table*}

\begin{table*}[ht]
\centering
\caption{Long Level Generated One Scene At a Time (64 wide). 
Using \texttt{MLM-regular0} (\url{https://huggingface.co/schrum2/MarioDiffusion-MLM-regular0}), 
the GUI was used to generate $64 \times 16$ scenes with the designated prompts. The segments
were then combined into a single playable level. Each segment of width 64 has its own
caption adherence score, but the result of splitting each segment into four scenes of
width 16 and averaging those caption adherence scores is also shown. Note that it is not necessary for level widths to be multiples of 16, nor is it necessary for all segments in a level to have the same width.}
\label{tab:long_long_segment_level}
\setlength{\tabcolsep}{0pt} 
\begin{tabular}{|*{2}{>{\centering\arraybackslash}p{0.5\textwidth}|}}
\hline
floor with several gaps. a few platforms. one rectangular block cluster. one irregular block cluster. a few enemies. a few coin lines. many coins. one ascending staircase. a few loose blocks. &
floor with two gaps. ceiling with one gap. several platforms. a few rectangular block clusters. one irregular block cluster. a few enemies. two pipes. one coin line. two coins. one cannon. several question blocks. several loose blocks. \\
\hline
\includegraphics[width=0.5\textwidth]{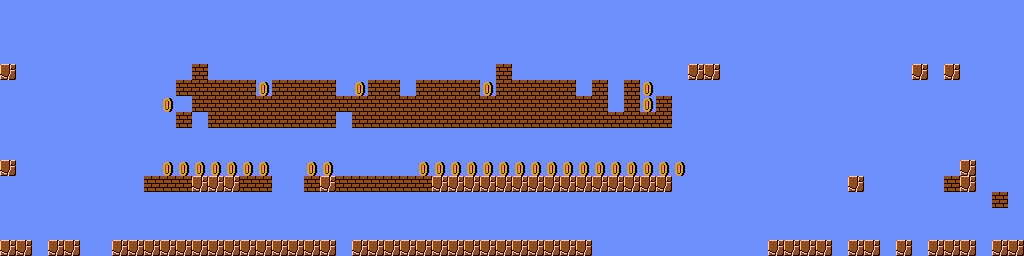} & 
\includegraphics[width=0.5\textwidth]{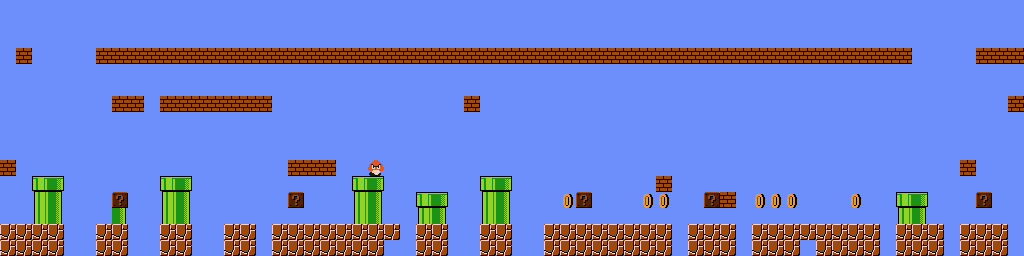} \\ 
\hline
0.638889 &
0.458333 \\
AVG: 0.479167 &
AVG: 0.280556 \\
\hline
\end{tabular}
\end{table*}

\begin{table*}[htbp]
\centering
\caption{Example scenes generated by models trained with regular captions. Each of these models is available on Hugging Face (Details here: \url{https://github.com/schrum2/MarioDiffusion/blob/main/MODELS.md}).
The first row shows the prompt used to generate the scene. The first five columns are real captions from the test set, and the next five are from the random test set of captions not present in the original data. Beneath each image is the resulting caption adherence score. These images are also available online: \url{https://people.southwestern.edu/~schrum2/mario.html}.}
\label{tab:regular_generated}
\setlength{\tabcolsep}{0pt} 
\begin{tabular}{|*{10}{>{\centering\arraybackslash}p{0.1\textwidth}|}}
\hline
full floor. one enemy. a few question blocks. one platform. one pipe. &
floor with one gap. one descending staircase. one pipe. one irregular block cluster. & 
full floor. full ceiling. one enemy. one coin. one irregular block cluster. a few towers. a few loose blocks. & 
floor with one gap. a few enemies. one cannon. one tower. & 
full floor. a few enemies. a few question blocks. one platform. one upside down pipe. two loose blocks. & 
a few coin lines. one irregular block cluster. a few enemies. several coins. two ascending staircases. one question block. one rectangular block cluster. two cannons. & 
floor with several gaps. two pipes. two enemies. one descending staircase. two towers. two upside down pipes. & 
full floor. one descending staircase. one loose block. a few upside down pipes. full ceiling. two coins. one enemy. & 
several platforms. two rectangular block clusters. one pipe. a few upside down pipes. & 
floor with several gaps. one tower. \\
\hline
\multicolumn{10}{|c|}{\texttt{MLM-regular0}} \\
\hline
\includegraphics[width=0.1\textwidth]{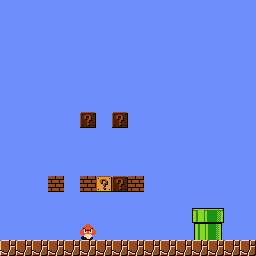} & 
\includegraphics[width=0.1\textwidth]{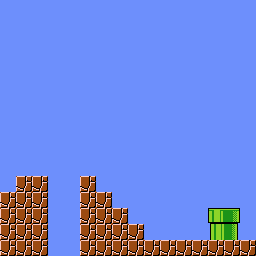} & 
\includegraphics[width=0.1\textwidth]{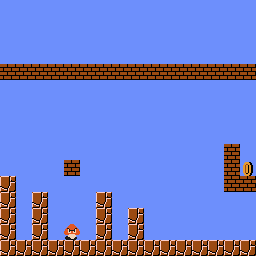} & 
\includegraphics[width=0.1\textwidth]{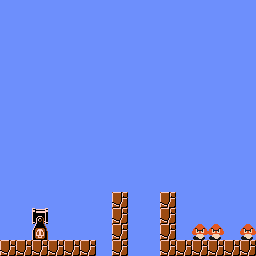} & 
\includegraphics[width=0.1\textwidth]{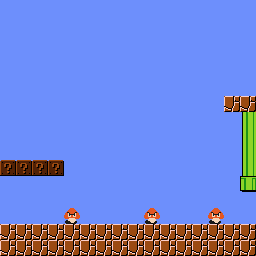} & 
\includegraphics[width=0.1\textwidth]{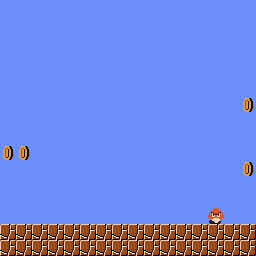} & 
\includegraphics[width=0.1\textwidth]{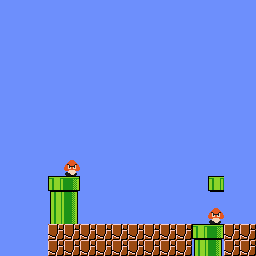} & 
\includegraphics[width=0.1\textwidth]{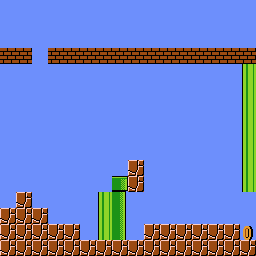} & 
\includegraphics[width=0.1\textwidth]{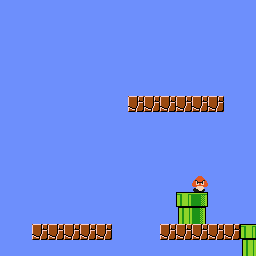} & 
\includegraphics[width=0.1\textwidth]{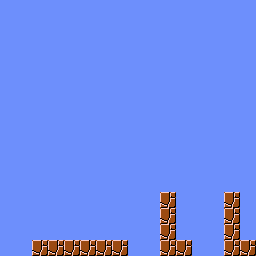} \\
\hline
0.88888889 & 
1.0 & 
1.0 & 
0.97222222 & 
1.0 & 
0.26388889 & 
0.51388889 & 
0.35277778 & 
0.625 & 
0.97222222 \\
\hline
\multicolumn{10}{|c|}{\texttt{MiniLM-single-regular0}} \\
\hline
\includegraphics[width=0.1\textwidth]{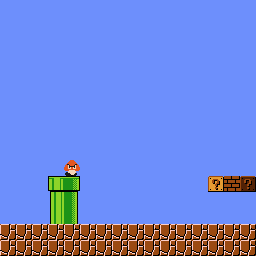} & 
\includegraphics[width=0.1\textwidth]{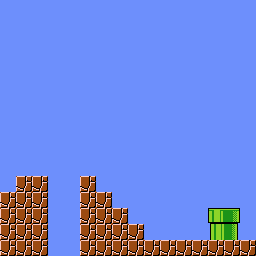} & 
\includegraphics[width=0.1\textwidth]{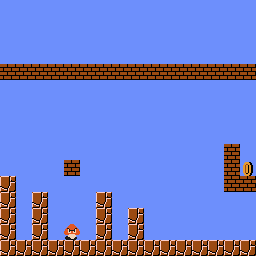} & 
\includegraphics[width=0.1\textwidth]{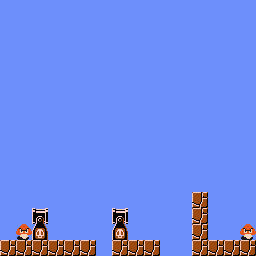} & 
\includegraphics[width=0.1\textwidth]{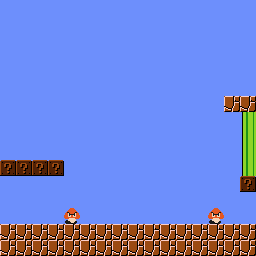} & 
\includegraphics[width=0.1\textwidth]{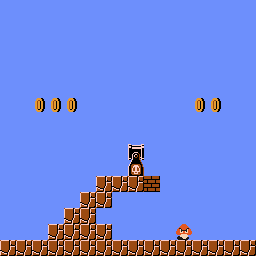} & 
\includegraphics[width=0.1\textwidth]{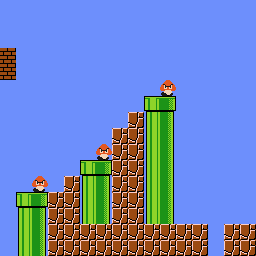} & 
\includegraphics[width=0.1\textwidth]{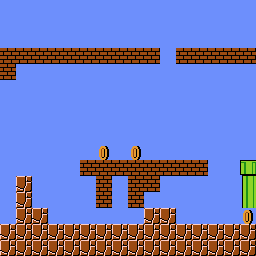} & 
\includegraphics[width=0.1\textwidth]{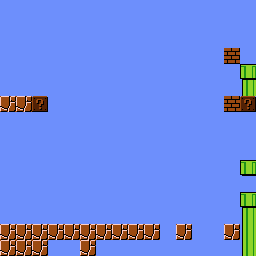} & 
\includegraphics[width=0.1\textwidth]{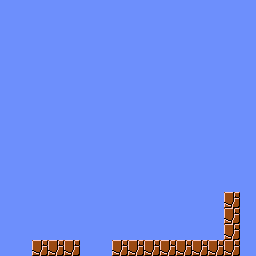} \\
\hline
0.98611111 & 
1.0 & 
1.0 & 
0.95833333 & 
0.75 & 
0.48611111 & 
0.38888889 & 
0.13055556 & 
0.5 & 
0.98611111 \\
\hline
\multicolumn{10}{|c|}{\texttt{MiniLM-multiple-regular0}} \\
\hline
\includegraphics[width=0.1\textwidth]{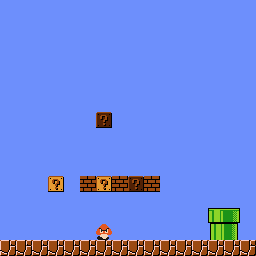} & 
\includegraphics[width=0.1\textwidth]{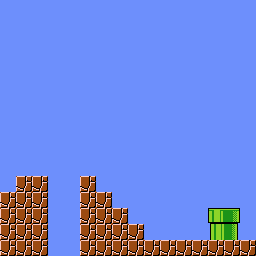} & 
\includegraphics[width=0.1\textwidth]{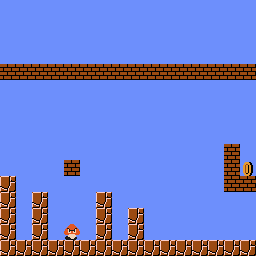} & 
\includegraphics[width=0.1\textwidth]{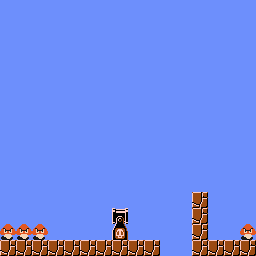} & 
\includegraphics[width=0.1\textwidth]{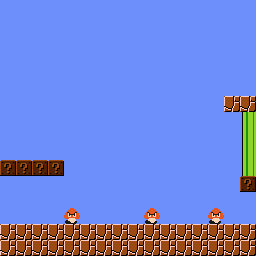} & 
\includegraphics[width=0.1\textwidth]{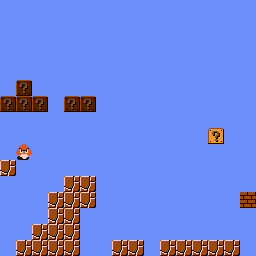} & 
\includegraphics[width=0.1\textwidth]{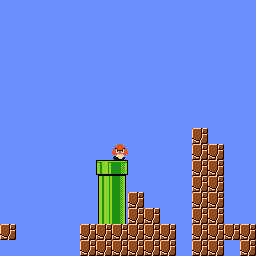} & 
\includegraphics[width=0.1\textwidth]{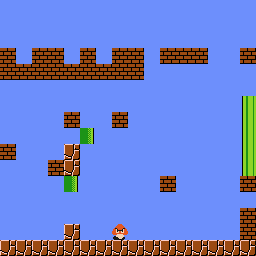} & 
\includegraphics[width=0.1\textwidth]{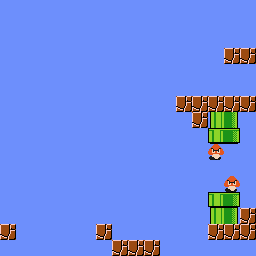} & 
\includegraphics[width=0.1\textwidth]{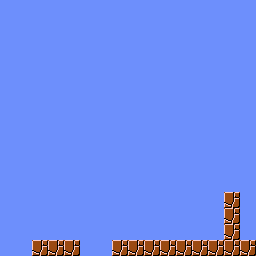} \\
\hline
1.0 & 
1.0 & 
1.0 & 
1.0 & 
0.76388889 & 
0.02777778 & 
0.40277778 & 
0.22777778 & 
0.375 & 
0.97222222 \\
\hline
\multicolumn{10}{|c|}{\texttt{GTE-single-regular0}} \\
\hline
\includegraphics[width=0.1\textwidth]{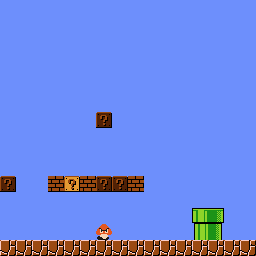} & 
\includegraphics[width=0.1\textwidth]{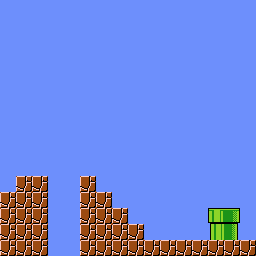} & 
\includegraphics[width=0.1\textwidth]{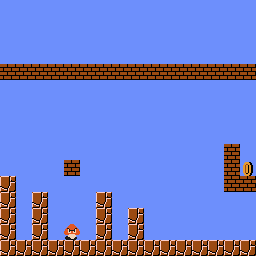} & 
\includegraphics[width=0.1\textwidth]{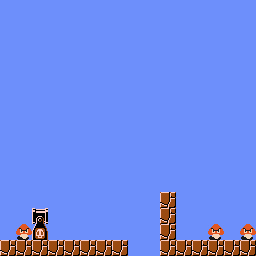} & 
\includegraphics[width=0.1\textwidth]{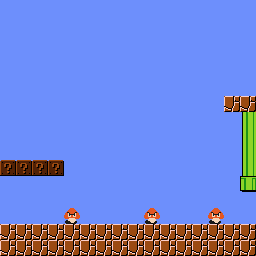} & 
\includegraphics[width=0.1\textwidth]{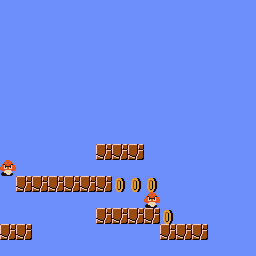} & 
\includegraphics[width=0.1\textwidth]{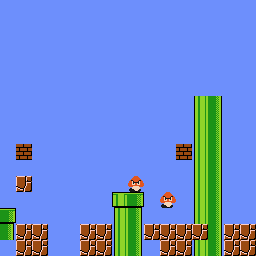} & 
\includegraphics[width=0.1\textwidth]{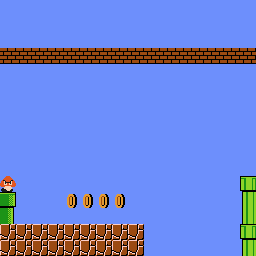} & 
\includegraphics[width=0.1\textwidth]{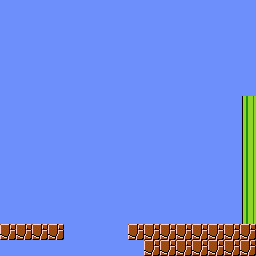} & 
\includegraphics[width=0.1\textwidth]{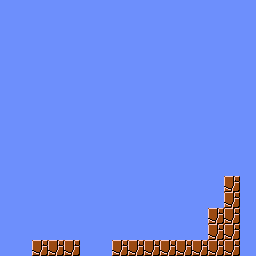} \\
\hline
0.98611111 & 
1.0 & 
1.0 & 
1.0 & 
1.0 & 
0.27777778 & 
0.31944444 & 
0.38055556 & 
0.5 & 
0.98611111 \\
\hline
\multicolumn{10}{|c|}{\texttt{GTE-multiple-regular0}} \\
\hline
\includegraphics[width=0.1\textwidth]{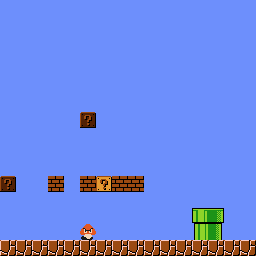} & 
\includegraphics[width=0.1\textwidth]{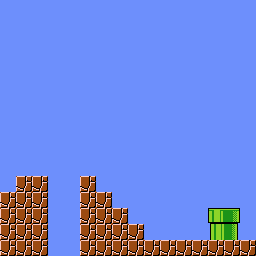} & 
\includegraphics[width=0.1\textwidth]{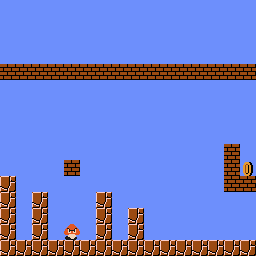} & 
\includegraphics[width=0.1\textwidth]{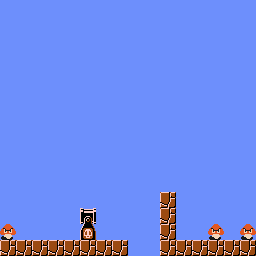} & 
\includegraphics[width=0.1\textwidth]{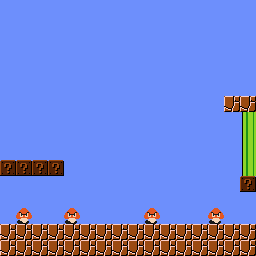} & 
\includegraphics[width=0.1\textwidth]{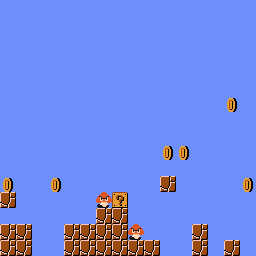} & 
\includegraphics[width=0.1\textwidth]{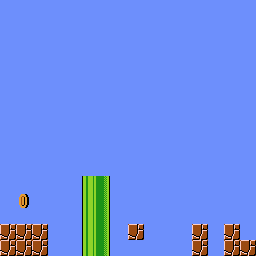} & 
\includegraphics[width=0.1\textwidth]{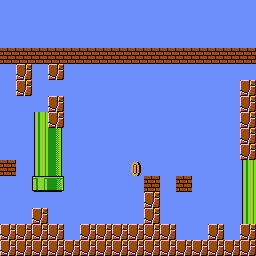} & 
\includegraphics[width=0.1\textwidth]{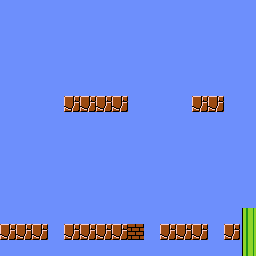} & 
\includegraphics[width=0.1\textwidth]{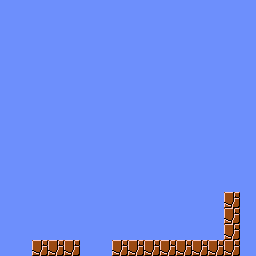} \\
\hline
0.88888889 & 
1.0 & 
1.0 & 
1.0 & 
0.76388889 & 
0.38888889 & 
$-0.0138889$ & 
0.375 & 
0.33333333 & 
0.98611111 \\
\hline
\end{tabular}
\end{table*}

\begin{table*}[htbp]
\centering
\caption{Example scenes generated by models trained with absence captions. Each of these models is available on Hugging Face (Details here: \url{https://github.com/schrum2/MarioDiffusion/blob/main/MODELS.md}).
The first row shows the regular prompt that the actual input prompt is based on. Phrases for absent concepts are added automatically. The first five columns are real captions from the test set, and the next five are from the random test set of captions not present in the original data. Beneath each image is the resulting caption adherence score. These images are also available online: \url{https://people.southwestern.edu/~schrum2/mario.html}.}
\label{tab:absent_generated}
\setlength{\tabcolsep}{0pt} 
\begin{tabular}{|*{10}{>{\centering\arraybackslash}p{0.1\textwidth}|}}
\hline
full floor. one enemy. a few question blocks. one platform. one pipe. &
floor with one gap. one descending staircase. one pipe. one irregular block cluster. & 
full floor. full ceiling. one enemy. one coin. one irregular block cluster. a few towers. a few loose blocks. & 
floor with one gap. a few enemies. one cannon. one tower. & 
full floor. a few enemies. a few question blocks. one platform. one upside down pipe. two loose blocks. & 
a few coin lines. one irregular block cluster. a few enemies. several coins. two ascending staircases. one question block. one rectangular block cluster. two cannons. & 
floor with several gaps. two pipes. two enemies. one descending staircase. two towers. two upside down pipes. & 
full floor. one descending staircase. one loose block. a few upside down pipes. full ceiling. two coins. one enemy. & 
several platforms. two rectangular block clusters. one pipe. a few upside down pipes. & 
floor with several gaps. one tower. \\
\hline
\multicolumn{10}{|c|}{\texttt{MLM-absence0}} \\
\hline
\includegraphics[width=0.1\textwidth]{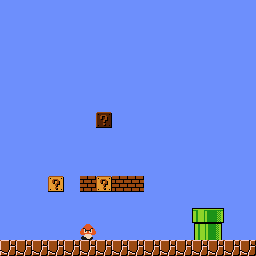} & 
\includegraphics[width=0.1\textwidth]{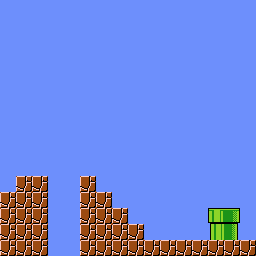} & 
\includegraphics[width=0.1\textwidth]{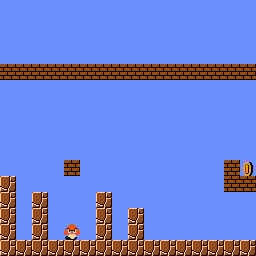} & 
\includegraphics[width=0.1\textwidth]{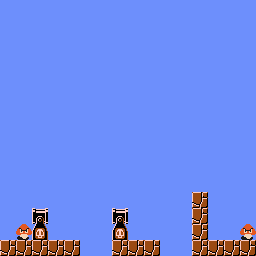} & 
\includegraphics[width=0.1\textwidth]{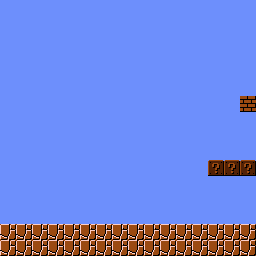} & 
\includegraphics[width=0.1\textwidth]{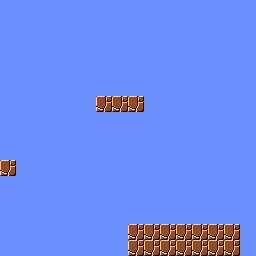} & 
\includegraphics[width=0.1\textwidth]{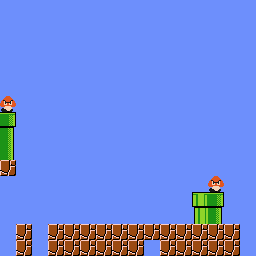} & 
\includegraphics[width=0.1\textwidth]{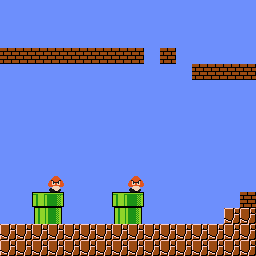} & 
\includegraphics[width=0.1\textwidth]{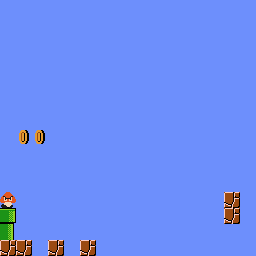} & 
\includegraphics[width=0.1\textwidth]{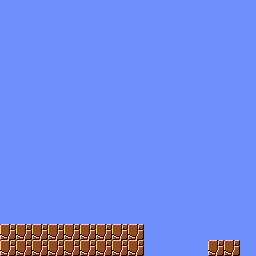} \\
\hline
1.0 & 
1.0 & 
1.0 & 
0.95833333 & 
0.76388889 & 
$-0.2222222$ & 
0.43055556 & 
0.15833333 & 
0.11111111 & 
0.75 \\
\hline
\multicolumn{10}{|c|}{\texttt{MiniLM-single-absence0}} \\
\hline
\includegraphics[width=0.1\textwidth]{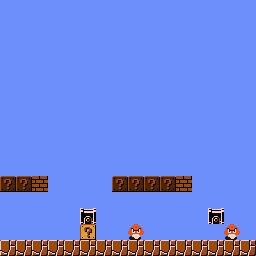} & 
\includegraphics[width=0.1\textwidth]{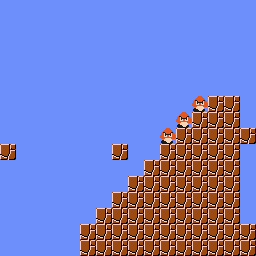} & 
\includegraphics[width=0.1\textwidth]{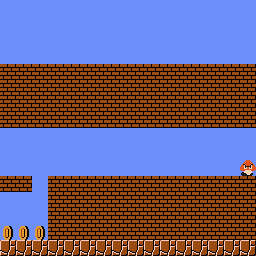} & 
\includegraphics[width=0.1\textwidth]{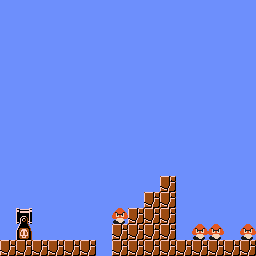} & 
\includegraphics[width=0.1\textwidth]{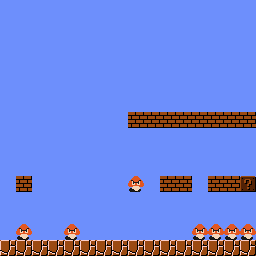} & 
\includegraphics[width=0.1\textwidth]{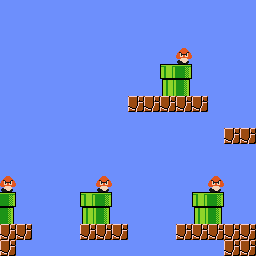} & 
\includegraphics[width=0.1\textwidth]{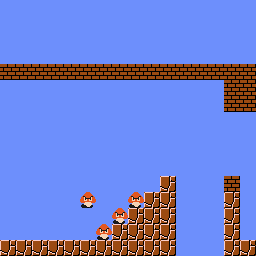} & 
\includegraphics[width=0.1\textwidth]{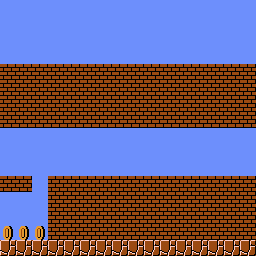} & 
\includegraphics[width=0.1\textwidth]{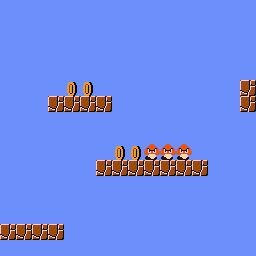} & 
\includegraphics[width=0.1\textwidth]{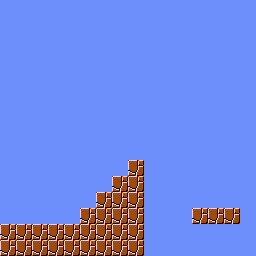} \\
\hline
0.73611111 & 
0.22222222 & 
0.30555556 & 
0.77777778 & 
0.80555556 & 
$-0.1388889$ & 
0.26388889 & 
0.20833333 & 
0.20833333 & 
0.625 \\
\hline
\multicolumn{10}{|c|}{\texttt{MiniLM-multiple-absence0}} \\
\hline
\includegraphics[width=0.1\textwidth]{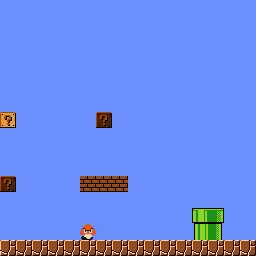} & 
\includegraphics[width=0.1\textwidth]{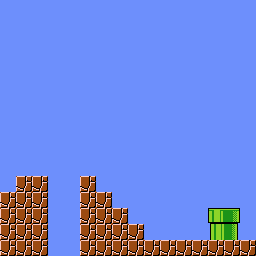} & 
\includegraphics[width=0.1\textwidth]{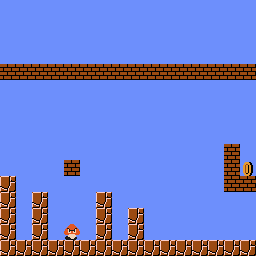} & 
\includegraphics[width=0.1\textwidth]{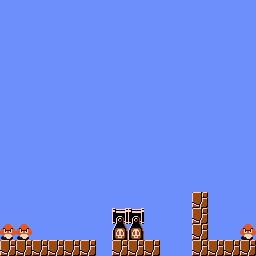} & 
\includegraphics[width=0.1\textwidth]{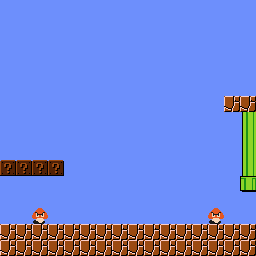} & 
\includegraphics[width=0.1\textwidth]{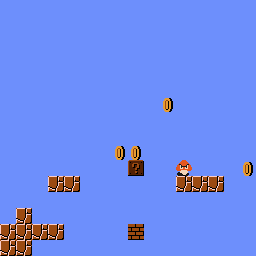} & 
\includegraphics[width=0.1\textwidth]{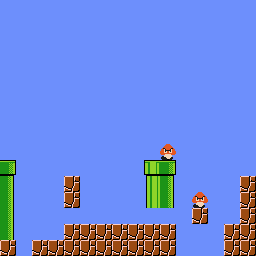} & 
\includegraphics[width=0.1\textwidth]{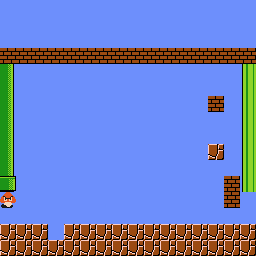} & 
\includegraphics[width=0.1\textwidth]{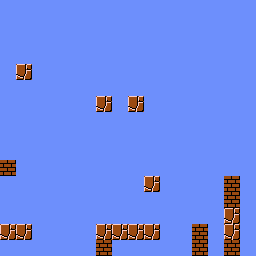} & 
\includegraphics[width=0.1\textwidth]{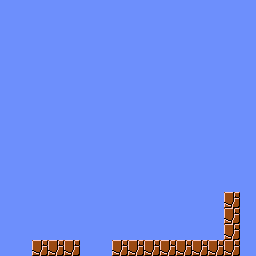} \\
\hline
1.0 & 
1.0 & 
1.0 & 
0.97222222 & 
0.98611111 & 
0.26388889 & 
0.51388889 & 
0.5 & 
0.30555556 & 
0.98611111 \\
\hline
\multicolumn{10}{|c|}{\texttt{GTE-single-absence0}} \\
\hline
\includegraphics[width=0.1\textwidth]{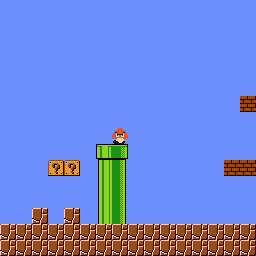} & 
\includegraphics[width=0.1\textwidth]{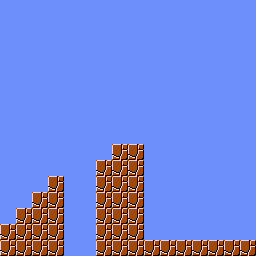} & 
\includegraphics[width=0.1\textwidth]{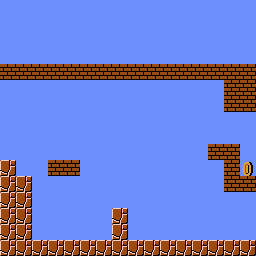} & 
\includegraphics[width=0.1\textwidth]{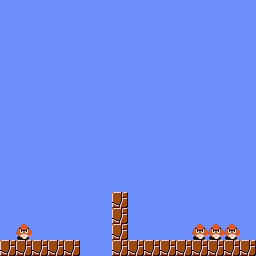} & 
\includegraphics[width=0.1\textwidth]{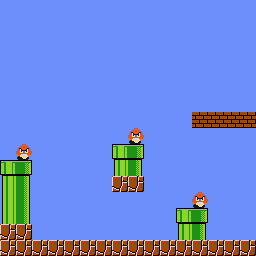} & 
\includegraphics[width=0.1\textwidth]{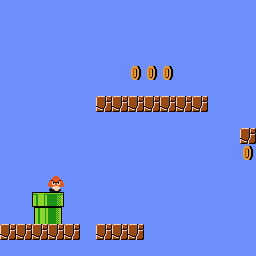} & 
\includegraphics[width=0.1\textwidth]{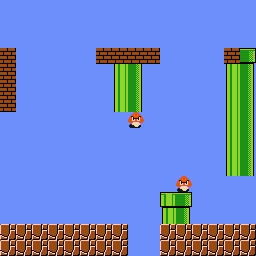} & 
\includegraphics[width=0.1\textwidth]{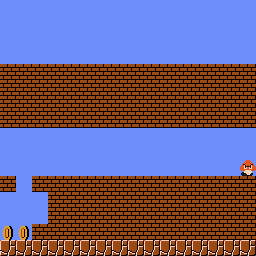} & 
\includegraphics[width=0.1\textwidth]{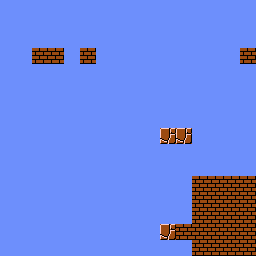} & 
\includegraphics[width=0.1\textwidth]{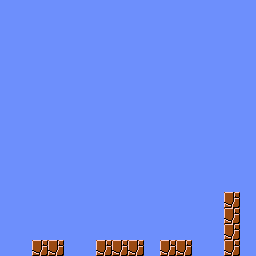} \\
\hline
0.86111111 & 
0.66666667 & 
0.625 & 
0.88888889 & 
0.65277778 & 
0.15277778 & 
0.27777778 & 
0.44444444 & 
0.52777778 & 
0.98611111 \\
\hline
\multicolumn{10}{|c|}{\texttt{GTE-multiple-absence0}} \\
\hline
\includegraphics[width=0.1\textwidth]{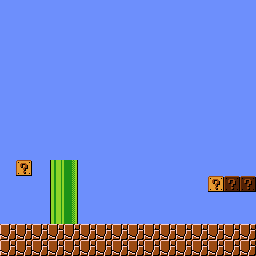} & 
\includegraphics[width=0.1\textwidth]{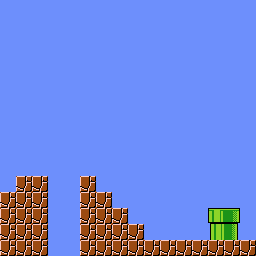} & 
\includegraphics[width=0.1\textwidth]{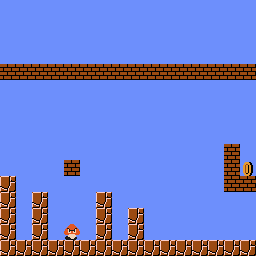} & 
\includegraphics[width=0.1\textwidth]{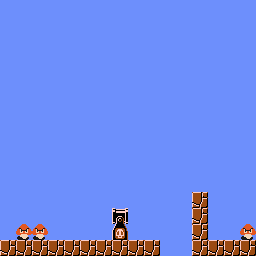} & 
\includegraphics[width=0.1\textwidth]{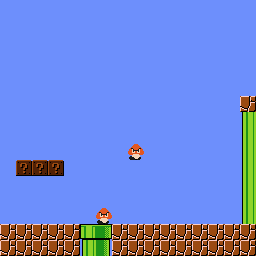} & 
\includegraphics[width=0.1\textwidth]{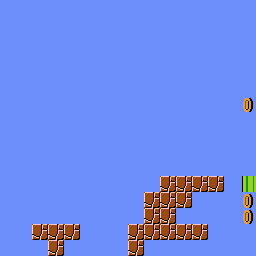} & 
\includegraphics[width=0.1\textwidth]{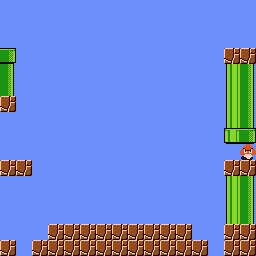} & 
\includegraphics[width=0.1\textwidth]{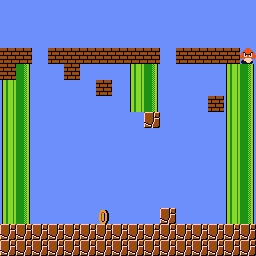} & 
\includegraphics[width=0.1\textwidth]{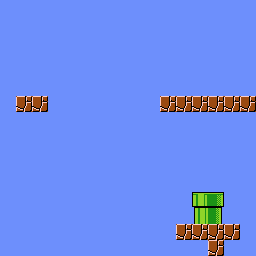} & 
\includegraphics[width=0.1\textwidth]{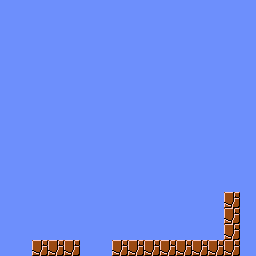} \\
\hline
0.66666667 & 
1.0 & 
1.0 & 
1.0 & 
0.63888889 & 
0.09722222 & 
0.26388889 & 
0.56111111 & 
0.73611111 & 
0.98611111 \\
\hline
\end{tabular}
\end{table*}

\begin{table*}[htbp]
\centering
\caption{Example scenes generated by models trained with negative captions. Each of these models is available on Hugging Face (Details here: \url{https://github.com/schrum2/MarioDiffusion/blob/main/MODELS.md}).
The first row shows the regular prompt. Phrases for absent concepts automatically create the corresponding negative prompt. The first five columns are real captions from the test set, and the next five are from the random test set of captions not present in the original data. Beneath each image is the resulting caption adherence score. These images are also available online: \url{https://people.southwestern.edu/~schrum2/mario.html}.}
\label{tab:negative_generated}
\setlength{\tabcolsep}{0pt} 
\begin{tabular}{|*{10}{>{\centering\arraybackslash}p{0.1\textwidth}|}}
\hline
full floor. one enemy. a few question blocks. one platform. one pipe. &
floor with one gap. one descending staircase. one pipe. one irregular block cluster. & 
full floor. full ceiling. one enemy. one coin. one irregular block cluster. a few towers. a few loose blocks. & 
floor with one gap. a few enemies. one cannon. one tower. & 
full floor. a few enemies. a few question blocks. one platform. one upside down pipe. two loose blocks. & 
a few coin lines. one irregular block cluster. a few enemies. several coins. two ascending staircases. one question block. one rectangular block cluster. two cannons. & 
floor with several gaps. two pipes. two enemies. one descending staircase. two towers. two upside down pipes. & 
full floor. one descending staircase. one loose block. a few upside down pipes. full ceiling. two coins. one enemy. & 
several platforms. two rectangular block clusters. one pipe. a few upside down pipes. & 
floor with several gaps. one tower. \\
\hline
\multicolumn{10}{|c|}{\texttt{MLM-negative0}} \\
\hline
\includegraphics[width=0.1\textwidth]{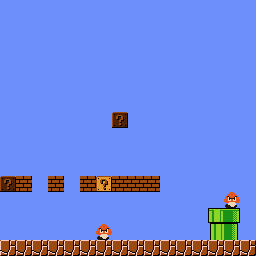} & 
\includegraphics[width=0.1\textwidth]{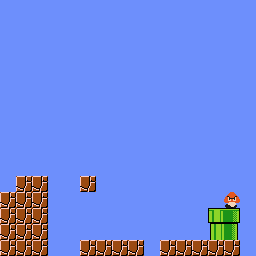} & 
\includegraphics[width=0.1\textwidth]{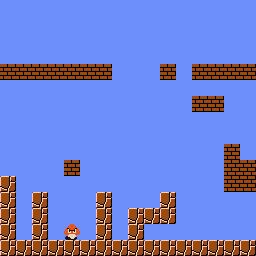} & 
\includegraphics[width=0.1\textwidth]{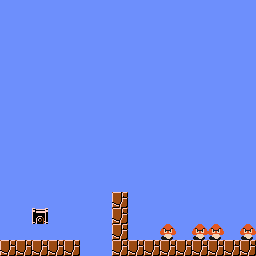} & 
\includegraphics[width=0.1\textwidth]{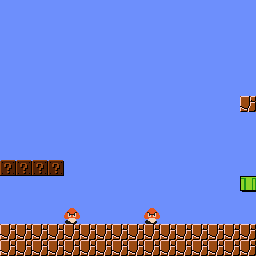} & 
\includegraphics[width=0.1\textwidth]{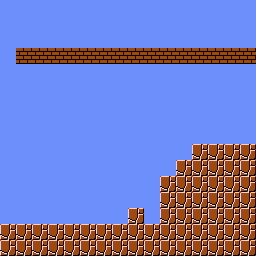} & 
\includegraphics[width=0.1\textwidth]{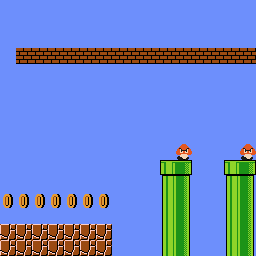} & 
\includegraphics[width=0.1\textwidth]{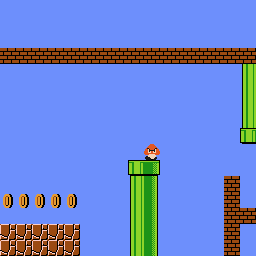} & 
\includegraphics[width=0.1\textwidth]{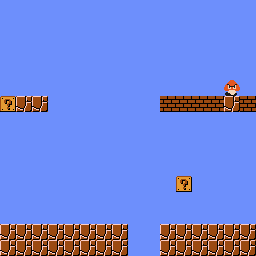} & 
\includegraphics[width=0.1\textwidth]{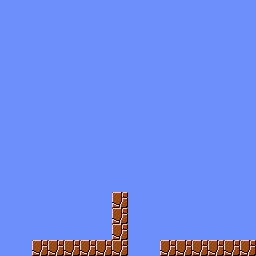} \\
\hline
0.86111111 & 
0.63888889 & 
0.68611111 & 
1.0 & 
0.75 & 
$-0.0138889$ & 
0.30555556 & 
0.33888889 & 
0.30555556 & 
0.97222222 \\
\hline
\multicolumn{10}{|c|}{\texttt{MiniLM-single-negative0}} \\
\hline
\includegraphics[width=0.1\textwidth]{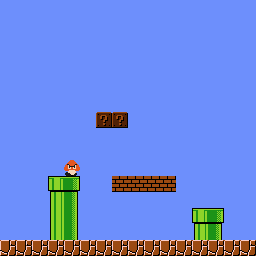} & 
\includegraphics[width=0.1\textwidth]{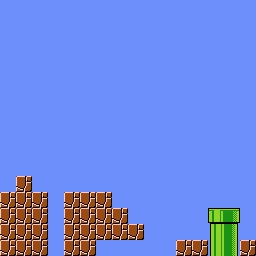} & 
\includegraphics[width=0.1\textwidth]{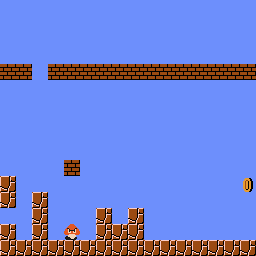} & 
\includegraphics[width=0.1\textwidth]{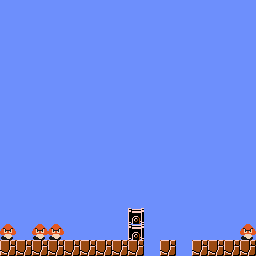} & 
\includegraphics[width=0.1\textwidth]{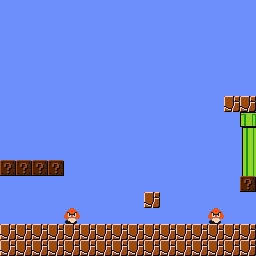} & 
\includegraphics[width=0.1\textwidth]{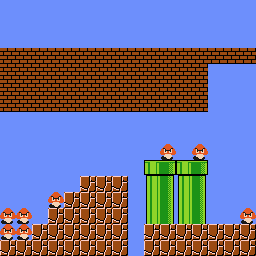} & 
\includegraphics[width=0.1\textwidth]{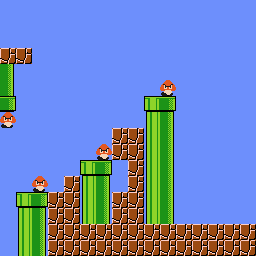} & 
\includegraphics[width=0.1\textwidth]{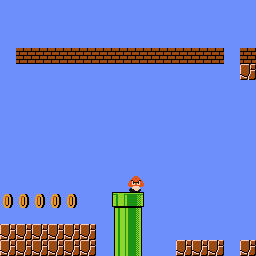} & 
\includegraphics[width=0.1\textwidth]{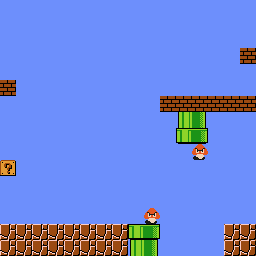} & 
\includegraphics[width=0.1\textwidth]{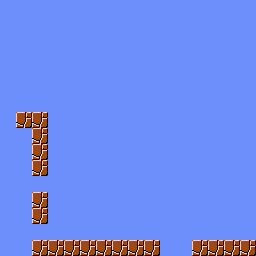} \\
\hline
0.95833333 & 
0.875 & 
0.92222222 & 
0.86111111 & 
0.73611111 & 
$-0.0416667$ & 
0.56944444 & 
0.31666667 & 
0.47222222 & 
0.63888889 \\
\hline
\multicolumn{10}{|c|}{\texttt{MiniLM-multiple-negative0}} \\
\hline
\includegraphics[width=0.1\textwidth]{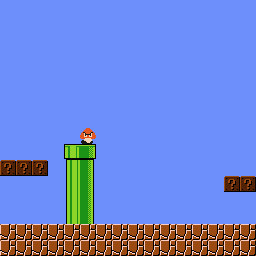} & 
\includegraphics[width=0.1\textwidth]{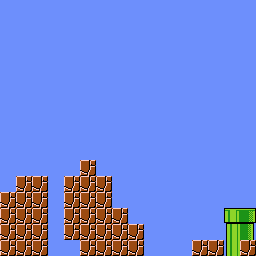} & 
\includegraphics[width=0.1\textwidth]{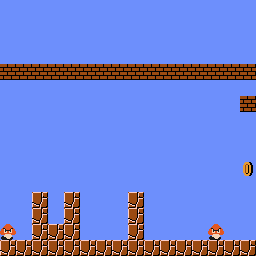} & 
\includegraphics[width=0.1\textwidth]{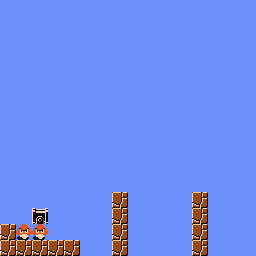} & 
\includegraphics[width=0.1\textwidth]{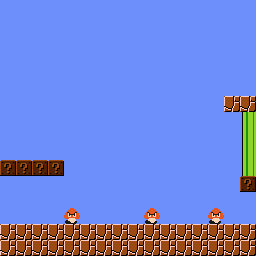} & 
\includegraphics[width=0.1\textwidth]{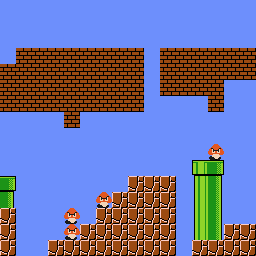} & 
\includegraphics[width=0.1\textwidth]{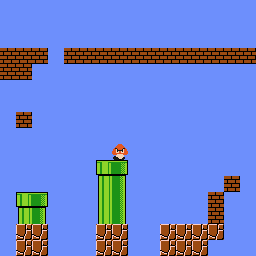} & 
\includegraphics[width=0.1\textwidth]{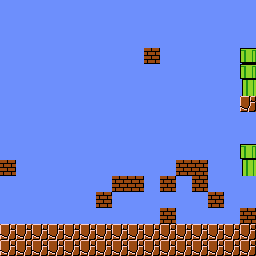} & 
\includegraphics[width=0.1\textwidth]{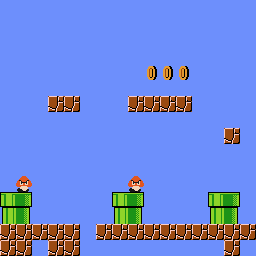} & 
\includegraphics[width=0.1\textwidth]{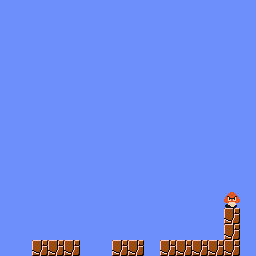} \\
\hline
0.97222222 & 
0.63888889 & 
0.93055556 & 
0.83333333 & 
0.76388889 & 
$-0.0416667$ & 
0.30555556 & 
0.06944444 & 
0.27777778 & 
0.65277778 \\
\hline
\multicolumn{10}{|c|}{\texttt{GTE-single-negative0}} \\
\hline
\includegraphics[width=0.1\textwidth]{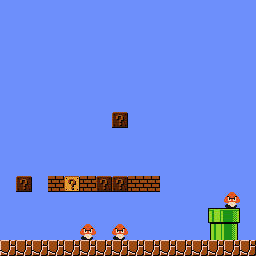} & 
\includegraphics[width=0.1\textwidth]{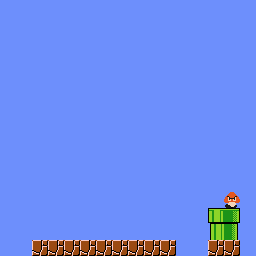} & 
\includegraphics[width=0.1\textwidth]{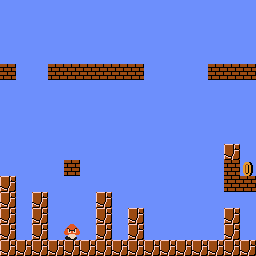} & 
\includegraphics[width=0.1\textwidth]{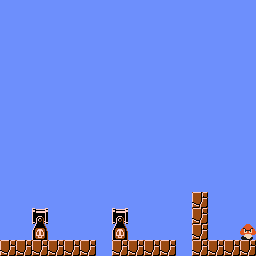} & 
\includegraphics[width=0.1\textwidth]{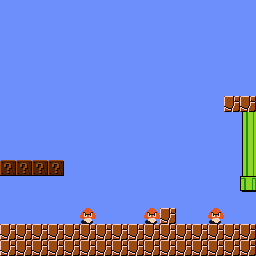} & 
\includegraphics[width=0.1\textwidth]{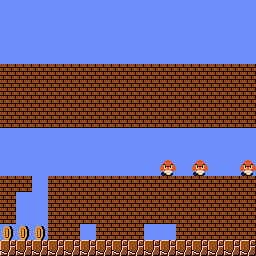} & 
\includegraphics[width=0.1\textwidth]{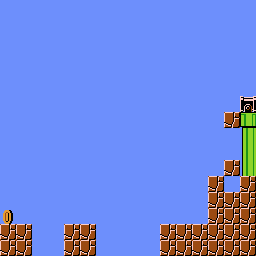} & 
\includegraphics[width=0.1\textwidth]{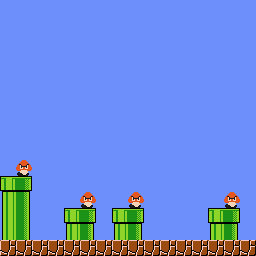} & 
\includegraphics[width=0.1\textwidth]{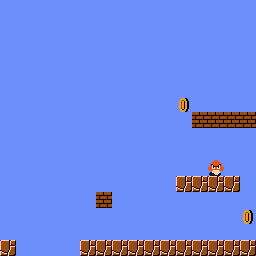} & 
\includegraphics[width=0.1\textwidth]{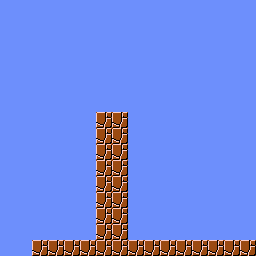} \\
\hline
0.95833333 & 
0.63888889 & 
0.93611111 & 
0.94444444 & 
0.98611111 & 
0.38888889 & 
0.06944444 & 
0.30555556 & 
0.19444444 & 
0.95833333 \\
\hline
\multicolumn{10}{|c|}{\texttt{GTE-multiple-negative0}} \\
\hline
\includegraphics[width=0.1\textwidth]{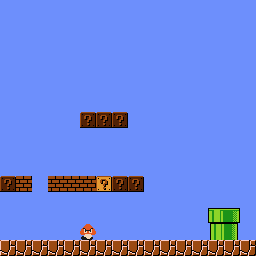} & 
\includegraphics[width=0.1\textwidth]{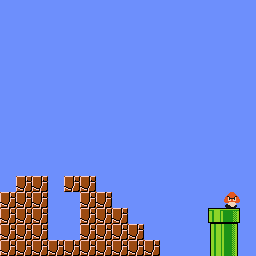} & 
\includegraphics[width=0.1\textwidth]{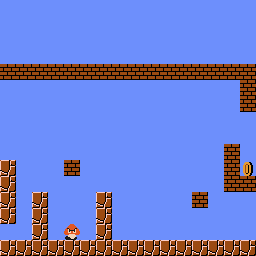} & 
\includegraphics[width=0.1\textwidth]{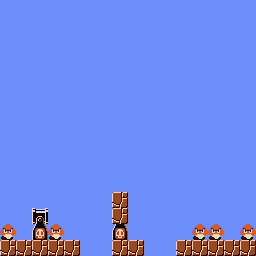} & 
\includegraphics[width=0.1\textwidth]{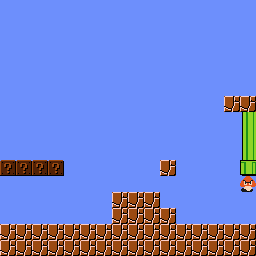} & 
\includegraphics[width=0.1\textwidth]{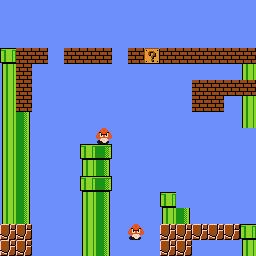} & 
\includegraphics[width=0.1\textwidth]{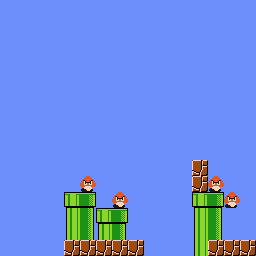} & 
\includegraphics[width=0.1\textwidth]{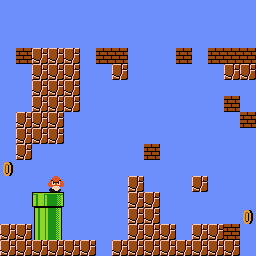} & 
\includegraphics[width=0.1\textwidth]{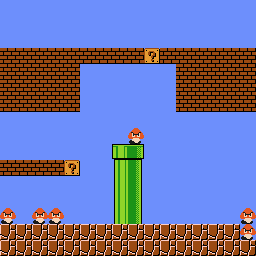} & 
\includegraphics[width=0.1\textwidth]{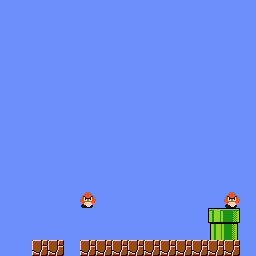} \\
\hline
0.95833333 & 
0.76388889 & 
0.875 & 
0.86111111 & 
0.73611111 & 
$-0.2361111$ & 
0.40277778 & 
0.30277778 & 
0.38888889 & 
0.65277778 \\
\hline
\end{tabular}
\end{table*}


\begin{thebibliography}{35}
\providecommand{\natexlab}[1]{#1}

\bibitem[{Awiszus, Schubert, and Rosenhahn(2020)}]{awiszus:aiide2020}
Awiszus, M.; Schubert, F.; and Rosenhahn, B. 2020.
\newblock TOAD-GAN: Coherent Style Level Generation from a Single Example.
\newblock In \emph{Artificial Intelligence and Interactive Digital
  Entertainment}. {AAAI}.

\bibitem[{Awiszus, Schubert, and Rosenhahn(2021)}]{awiszus:cog2021worldgan}
Awiszus, M.; Schubert, F.; and Rosenhahn, B. 2021.
\newblock World-GAN: a Generative Model for Minecraft Worlds.
\newblock In \emph{Conference on Games}, 1–8. IEEE.

\bibitem[{Bontrager et~al.(2018)Bontrager, Lin, Togelius, and
  Risi}]{bontrager2018deep}
Bontrager, P.; Lin, W.; Togelius, J.; and Risi, S. 2018.
\newblock Deep Interactive Evolution.
\newblock In \emph{European Conference on the Applications of Evolutionary
  Computation (EvoApplications)}.

\bibitem[{Dai et~al.(2024)Dai, Zhu, Li, Dai, and Wang}]{dai:aaai24}
Dai, S.; Zhu, X.; Li, N.; Dai, T.; and Wang, Z. 2024.
\newblock Procedural Level Generation with Diffusion Models from a Single
  Example.
\newblock \emph{Proceedings of the AAAI Conference on Artificial Intelligence},
  38(9): 10021--10029.

\bibitem[{Fontaine et~al.(2021)Fontaine, Hsu, Zhang, and
  Nikolaidis}]{fontaine:rss2021}
Fontaine, M.; Hsu, Y.-C.; Zhang, Y.; and Nikolaidis, S. 2021.
\newblock On the Importance of Environments for Human-Robot Coordination.
\newblock In \emph{Proceedings of Robotics: Science and Systems}.

\bibitem[{Goodfellow et~al.(2014)Goodfellow, Pouget-Abadie, Mirza, Xu,
  Warde-Farley, Ozair, Courville, and Bengio}]{goodfellow2014generative}
Goodfellow, I.; Pouget-Abadie, J.; Mirza, M.; Xu, B.; Warde-Farley, D.; Ozair,
  S.; Courville, A.; and Bengio, Y. 2014.
\newblock {Generative Adversarial Nets}.
\newblock In \emph{Neural Information Processing Systems}, 2672--2680.

\bibitem[{Gutierrez and Schrum(2020)}]{gutierrez2020zeldagan}
Gutierrez, J.; and Schrum, J. 2020.
\newblock {Generative Adversarial Network Rooms in Generative Graph Grammar
  Dungeons for The Legend of Zelda}.
\newblock In \emph{Congress on Evolutionary Computation}. IEEE.

\bibitem[{Guzdial, Liao, and Riedl(2018)}]{guzdial:aiideEXAG2018}
Guzdial, M.; Liao, N.; and Riedl, M. 2018.
\newblock Co-Creative Level Design via Machine Learning.
\newblock In \emph{Proceedings of the Experimental AI in Games (EXAG) Workshop
  at AIIDE}.

\bibitem[{Kingma and Welling(2014)}]{kingma:iclr13}
Kingma, D.~P.; and Welling, M. 2014.
\newblock Auto-Encoding Variational Bayes.
\newblock In \emph{International Conference on Learning Representations}.

\bibitem[{Kumaran, Mott, and Lester(2020)}]{kumaran:aiide2020}
Kumaran, V.; Mott, B.~W.; and Lester, J.~C. 2020.
\newblock Generating Game Levels for Multiple Distinct Games with a Common
  Latent Space.
\newblock In \emph{Artificial Intelligence and Interactive Digital
  Entertainment}. {AAAI}.

\bibitem[{Larsson, Font, and Alvarez(2022)}]{larsson:aiide2022}
Larsson, T.; Font, J.; and Alvarez, A. 2022.
\newblock Towards AI as a Creative Colleague in Game Level Design.
\newblock In \emph{Artificial Intelligence and Interactive Digital
  Entertainment}. {AAAI}.

\bibitem[{Lee and Simo-Serra(2023)}]{lee:mva23}
Lee, H.~J.; and Simo-Serra, E. 2023.
\newblock Using Unconditional Diffusion Models in Level Generation for Super
  Mario Bros.
\newblock In \emph{International Conference on Machine Vision and
  Applications}, 1--5.

\bibitem[{Merino et~al.(2023)Merino, Negri, Rajesh, Charity, and
  Togelius}]{merino:aaai23}
Merino, T.; Negri, R.; Rajesh, D.; Charity, M.; and Togelius, J. 2023.
\newblock The Five-Dollar Model: Generating Game Maps and Sprites From Sentence
  Embeddings.
\newblock In \emph{Artificial Intelligence and Interactive Digital
  Entertainment}. AAAI.

\bibitem[{Nie et~al.(2025)Nie, Middleton, Merino, Kanagaraja, Kumar, Zhuang,
  and Togelius}]{nie:aaai2025}
Nie, Y.; Middleton, M.; Merino, T.; Kanagaraja, N.; Kumar, A.; Zhuang, Z.; and
  Togelius, J. 2025.
\newblock Moonshine: Distilling Game Content Generators into Steerable
  Generative Models.
\newblock \emph{Proceedings of the AAAI Conference on Artificial Intelligence},
  39(13): 14344--14351.

\bibitem[{Radford et~al.(2019)Radford, Wu, Child, Luan, Amodei, and
  Sutskever}]{radford2019openAIGPT2}
Radford, A.; Wu, J.; Child, R.; Luan, D.; Amodei, D.; and Sutskever, I. 2019.
\newblock Language models are unsupervised multitask learners.
\newblock \emph{OpenAI Blog}.

\bibitem[{Rombach et~al.(2022)Rombach, Blattmann, Lorenz, Esser, and
  Ommer}]{rombach:cvpr2022SD}
Rombach, R.; Blattmann, A.; Lorenz, D.; Esser, P.; and Ommer, B. 2022.
\newblock { High-Resolution Image Synthesis with Latent Diffusion Models }.
\newblock In \emph{Computer Vision and Pattern Recognition}, 10674--10685.
  IEEE.

\bibitem[{Sarkar and Cooper(2021)}]{sarkar2021generating}
Sarkar, A.; and Cooper, S. 2021.
\newblock Generating and Blending Game Levels via Quality-Diversity in the
  Latent Space of a Variational Autoencoder.
\newblock In \emph{Proceedings of the Foundations of Digital Games}.

\bibitem[{Sarkar, Yang, and Cooper(2020)}]{sarkar2020conditional}
Sarkar, A.; Yang, Z.; and Cooper, S. 2020.
\newblock Conditional Level Generation and Game Blending.
\newblock In \emph{Proceedings of the Experimental AI in Games (EXAG) Workshop
  at AIIDE}.

\bibitem[{Schrum et~al.(2023)Schrum, Capps, Steckel, Volz, and
  Risi}]{schrum:tog2023}
Schrum, J.; Capps, B.; Steckel, K.; Volz, V.; and Risi, S. 2023.
\newblock Hybrid Encoding for Generating Large Scale Game Level Patterns With
  Local Variations.
\newblock \emph{IEEE Transactions on Games}, 15(1): 46--55.

\bibitem[{Schrum et~al.(2020)Schrum, Gutierrez, Volz, Liu, Lucas, and
  Risi}]{schrum2020interactive}
Schrum, J.; Gutierrez, J.; Volz, V.; Liu, J.; Lucas, S.; and Risi, S. 2020.
\newblock Interactive Evolution and Exploration Within Latent Level-Design
  Space of Generative Adversarial Networks.
\newblock In \emph{Genetic and Evolutionary Computation Conference}. ACM.

\bibitem[{Schrum, Volz, and Risi(2020)}]{schrum:gecco2020cppn2gan}
Schrum, J.; Volz, V.; and Risi, S. 2020.
\newblock CPPN2GAN: Combining Compositional Pattern Producing Networks and GANs
  for Large-scale Pattern Generation.
\newblock In \emph{Genetic and Evolutionary Computation Conference}. ACM.

\bibitem[{Shaker, Togelius, and Nelson(2016)}]{shaker2016procedural}
Shaker, N.; Togelius, J.; and Nelson, M.~J. 2016.
\newblock \emph{Procedural Content Generation in Games}.
\newblock Springer.

\bibitem[{Sudhakaran et~al.(2023)Sudhakaran, Gonz\'{a}lez-Duque, Freiberger,
  Glanois, Najarro, and Risi}]{sudhakaran:nips23}
Sudhakaran, S.; Gonz\'{a}lez-Duque, M.; Freiberger, M.; Glanois, C.; Najarro,
  E.; and Risi, S. 2023.
\newblock MarioGPT: Open-Ended Text2Level Generation Through Large Language
  Models.
\newblock In \emph{Neural Information Processing Systems}.

\bibitem[{Summerville and Mateas(2016)}]{summerville2016super}
Summerville, A.; and Mateas, M. 2016.
\newblock Super Mario as a String: Platformer Level Generation via LSTMs.
\newblock In \emph{1st International Joint Conference of DiGRA and FDG}.

\bibitem[{Summerville et~al.(2018)Summerville, Snodgrass, Guzdial,
  Holmg{\aa}rd, Hoover, Isaksen, Nealen, and Togelius}]{summerville:tog2018}
Summerville, A.; Snodgrass, S.; Guzdial, M.; Holmg{\aa}rd, C.; Hoover, A.~K.;
  Isaksen, A.; Nealen, A.; and Togelius, J. 2018.
\newblock Procedural Content Generation via Machine Learning (PCGML).
\newblock \emph{IEEE Transactions on Games}, 10(3): 257--270.

\bibitem[{Summerville et~al.(2016)Summerville, Snodgrass, Mateas, and
  Onta{\~n}{\'o}n}]{summerville:vglc2016}
Summerville, A.~J.; Snodgrass, S.; Mateas, M.; and Onta{\~n}{\'o}n, S. 2016.
\newblock {The VGLC: The Video Game Level Corpus}.
\newblock In \emph{Procedural Content Generation in Games}. ACM.

\bibitem[{Thakkar et~al.(2019)Thakkar, Cao, Wang, Choi, and
  Togelius}]{thakkar:cog19}
Thakkar, S.; Cao, C.; Wang, L.; Choi, T.~J.; and Togelius, J. 2019.
\newblock Autoencoder and Evolutionary Algorithm for Level Generation in Lode
  Runner.
\newblock In \emph{Conference on Games}, 1–4. IEEE.

\bibitem[{Todd et~al.(2023)Todd, Earle, Nasir, Green, and
  Togelius}]{todd:fdg23}
Todd, G.; Earle, S.; Nasir, M.~U.; Green, M.~C.; and Togelius, J. 2023.
\newblock Level Generation Through Large Language Models.
\newblock In \emph{Foundations of Digital Games}. ACM.

\bibitem[{Togelius, Karakovskiy, and
  Baumgarten(2010)}]{togelius:cig2010:marioAI}
Togelius, J.; Karakovskiy, S.; and Baumgarten, R. 2010.
\newblock {The 2009 Mario AI Competition}.
\newblock \emph{Congress on Evolutionary Computation}, 1--8.

\bibitem[{Valevski et~al.(2024)Valevski, Leviathan, Arar, and
  Fruchter}]{valevski2024diffusionmodelsrealtimegame}
Valevski, D.; Leviathan, Y.; Arar, M.; and Fruchter, S. 2024.
\newblock Diffusion Models Are Real-Time Game Engines.
\newblock arXiv:2408.14837.

\bibitem[{{Virtuals Protocol}(2024)}]{virtuals2024videogame}
{Virtuals Protocol}. 2024.
\newblock Video Game Generation: A Practical Study Using Mario.
\newblock Preprint.

\bibitem[{Volz et~al.(2018)Volz, Schrum, Liu, Lucas, Smith, and
  Risi}]{volz:gecco2018}
Volz, V.; Schrum, J.; Liu, J.; Lucas, S.~M.; Smith, A.~M.; and Risi, S. 2018.
\newblock {Evolving Mario Levels in the Latent Space of a Deep Convolutional
  Generative Adversarial Network}.
\newblock In \emph{Genetic and Evolutionary Computation Conference}. ACM.

\bibitem[{{{\v{S}}}osvald et~al.(2021){{\v{S}}}osvald, T{\"o}pfer, Holan,
  {{\v{C}}}ern{\'y}, and Gemrot}]{sosvald:ictai2021marioastar}
{{\v{S}}}osvald, D.; T{\"o}pfer, M.; Holan, J.; {{\v{C}}}ern{\'y}, V.; and
  Gemrot, J. 2021.
\newblock Super Mario A-Star Agent Revisited.
\newblock In \emph{International Conference on Tools with Artificial
  Intelligence}, 1008--1012. IEEE.

\bibitem[{Yang et~al.(2023)Yang, Zhang, Song, Hong, Xu, Zhao, Zhang, Cui, and
  Yang}]{yang2023diffusionsurvey}
Yang, L.; Zhang, Z.; Song, Y.; Hong, S.; Xu, R.; Zhao, Y.; Zhang, W.; Cui, B.;
  and Yang, M.-H. 2023.
\newblock Diffusion Models: A Comprehensive Survey of Methods and Applications.
\newblock \emph{ACM Computing Surveys}, 56(4).

\bibitem[{Zhang et~al.(2024)Zhang, Zhang, Long, Xie, Dai, Tang, Lin, Yang, Xie,
  Huang, Zhang, Li, and Zhang}]{zhang-etal-2024-mgte}
Zhang, X.; Zhang, Y.; Long, D.; Xie, W.; Dai, Z.; Tang, J.; Lin, H.; Yang, B.;
  Xie, P.; Huang, F.; Zhang, M.; Li, W.; and Zhang, M. 2024.
\newblock {mGTE}: Generalized Long-Context Text Representation and Reranking
  Models for Multilingual Text Retrieval.
\newblock In \emph{Empirical Methods in Natural Language Processing: Industry
  Track}, 1393--1412. Association for Computational Linguistics.

\end{thebibliography}
\end{document}